%% file: lifelong_generative_modeling_submitted.tex
\documentclass{article} 
\usepackage{iclr2019_conference,times}

\iclrfinaltrue

\input{math_commands.tex}

\usepackage{lineno,hyperref}

\usepackage{graphicx}

\usepackage{url}
\usepackage{amssymb}
\usepackage{amsmath}
\usepackage{enumitem}
\usepackage{gensymb}
\usepackage{stmaryrd}
\usepackage{fancyhdr}
\usepackage{makecell}
\usepackage{adjustbox}
\usepackage{pifont}%
\usepackage{bm}
\usepackage{bbm} 
\usepackage{relsize}
\usepackage{bmpsize}
\usepackage{graphicx} 
\usepackage{colonequals}
\newcommand*{\logeq}{\Leftrightarrow}
\usepackage[font=small,labelfont=bf,tableposition=top]{caption}
\DeclareCaptionLabelFormat{andtable}{#1~#2  \&  \tablename~\thetable}
\usepackage{mathtools}
\RequirePackage{algorithm}
\RequirePackage{algorithmic}

\newcounter{captionedequationset} 
\newdimen\captionlength
\newcommand{\captionedequationset}[1]{
    \refstepcounter{captionedequationset}
    \setlength{\captionlength}{\widthof{#1}} %
    \addtolength{\captionlength}{\widthof{Equation ~\thecaptionedequationset: }}
    \ifthenelse{\lengthtest{\captionlength < \linewidth }} %
    {\begin{center}
            Equation ~\thecaptionedequationset: #1
        \end{center}}
    { \begin{flushleft}
        Equation ~\thecaptionedequationset: #1 %
        \end{flushleft}}}

\newtheorem{lemma}{Lemma}
\newtheorem{proof}{Proof}

\usepackage[turnoff]{notes}

\newcommand{\vect}[1]{\bm{\mathrm{#1}}}

\title{Lifelong Generative Modeling}

\newcommand*\samethanks[1][\value{footnote}]{\footnotemark[#1]}

\author{
  Jason Ramapuram \thanks{University of Geneva, Switzerland}\ \ \thanks{Haute école
    de gestion de Genève, HES-SO, Switzerland}\\
  \texttt{Jason.Ramapuram@etu.unige.ch} \\
  \AND
  Magda Gregorova \samethanks[1] \ \samethanks[2]\\
  \texttt{magda.gregorova@hesge.ch} \\
  \And
  Alexandros Kalousis \samethanks[1] \ \samethanks[2]\\
  \texttt{Alexandros.Kalousis@hesge.ch} \\
}

\begin{document}
\maketitle

\vspace{-0.2in}
\begin{abstract}
Lifelong learning 
is the problem of learning multiple consecutive tasks in a
sequential manner, where knowledge gained from previous tasks is retained
and used to aid future learning over the lifetime of the learner. It is essential towards the development of
intelligent machines that can adapt to their surroundings. In this work we focus
on a lifelong learning approach to unsupervised generative modeling, where we continuously
incorporate newly observed distributions into a learned model. We do
so through a student-teacher Variational Autoencoder 
architecture which allows us to learn and preserve
all the distributions seen so far, without the need to retain the past
data nor the past models. Through the introduction of a novel cross-model
regularizer, inspired by a Bayesian update rule, the student model leverages the information learned by the teacher,
which acts as a probabilistic knowledge store. The regularizer reduces the effect of catastrophic interference that
appears when we learn over sequences of distributions. We
validate our model's performance on sequential variants of MNIST,
FashionMNIST, PermutedMNIST, SVHN and Celeb-A and demonstrate
that our model mitigates the effects of catastrophic interference
faced by neural networks in sequential learning scenarios.
Our code is available at : \url{https://github.com/jramapuram/LifelongVAE_pytorch}.
\end{abstract}

\input{introduction}
\input{related}
\input{background}
\input{model}
\input{related_plus_plus}
\input{experiments}
\bibliographystyle{ieee}
\bibliography{lifelong_generative_modeling}

\newpage
\onecolumn
\input{appendix}

\end{document}

%% file: math_commands.tex
\usepackage{amsmath,amsfonts,bm}

\def\eqref#1{equation~\ref{#1}}

\def\1{\bm{1}}

\DeclareMathAlphabet{\mathsfit}{\encodingdefault}{\sfdefault}{m}{sl}
\SetMathAlphabet{\mathsfit}{bold}{\encodingdefault}{\sfdefault}{bx}{n}



%% file: introduction.tex
\section{Introduction}

Machine learning is the process of approximating unknown functions through
the observation of typically noisy data samples. Supervised
learning approximates these functions by learning a mapping from inputs to a
predefined set of outputs such as categorical class labels (classification) or continuous targets (regression).
\note[Replaced-18]{Unsupervised learning on the other hand attempts to learn an estimate of the inputs
using the inputs alone.} Unsupervised learning seeks to uncover structure and patterns from the
input data without any supervision. Examples of this learning paradigm
include density estimation and clustering methods.
Both learning paradigms make assumptions that restrict the set of plausible
solutions. These assumptions are referred to as
\emph{hypothesis spaces}, \emph{biases} or \emph{priors} and aid the
model in favoring one solution over another
\cite{mitchell1980need,vapnik2006estimation}. For example, the use of convolutions
\cite{lecun1995convolutional} to process images favors local
structure; recurrent models
\cite{Jordan:1990:ADP:104134.104148,hochreiter1997long} exploit
sequential dependencies and graph neural networks
\cite{scarselli2008graph,kipf2016semi} assume that the underlying data
can be modeled accurately as a graph.

Current state of the art machine-learning models typically focus on
learning a single model for a single task, such as image classification
\cite{liu2018progressive,szegedy2016inception,he2016deep,DBLP:journals/corr/SimonyanZ14a,krizhevsky2012imagenet},
image generation
\cite{DBLP:journals/corr/abs-1906-00446,DBLP:conf/iclr/BrockDS19,kingma2014,goodfellow2014generative},
natural language question answering \cite{devlin2019bert,radford2019language} or single game playing
\cite{vinyals2019alphastar,silver2016mastering}. In contrast, humans
experience a sequence of learning tasks over their lifetimes, and
are able to leverage previous learning experiences to rapidly learn
new tasks. Consider learning how to ride a motorbike after learning
to ride a bicycle: the task is drastically simplified through the use
of prior learning experience. Studies \cite{ahn1993psychological,ahn1987schema,lake2015human} in psychology  have shown that humans are
able to generalize to new concepts in a rapid manner, given only a
handful of samples. \cite{lake2015human} demonstrates that humans can
classify and generate new concepts of two wheel vehicles given just a single
related sample. This contrasts the state of the art machine learning
models described above which use hundreds of thousands of samples and fail to generalize to
slight variations of the original task \cite{cobbe2019quantifying}.


Lifelong learning \cite{thrun1995lifelong,thrun1995lifelong2} argues
for the need to consider learning over task sequences, 
where learned task representations and models are \emph{stored} over the \emph{entire lifetime} of the
learner and can be used to aid current and future learning.
This form of learning allows for the transfer
of previously learned models and representations and can reduce the sample
complexity of the current learning problem
\cite{thrun1995lifelong}. In this work we restrict ourselves to a
subset of the broad lifelong learning paradigm; 
rather than focus on the supervised lifelong learning scenario as most
state of the art methods, our work is one of the first to tackle the
more challenging problem of deep lifelong
unsupervised learning. We also identity and relax crucial limitations of prior work in life-long learning
that requires the storage of previous models and training data,
allowing us to operate in a more realistic learning scenario.

%% file: related.tex
\vspace{-0.in}
\section{Related Work}\label{rel_sec}
\vspace{-0.1in}

The idea of learning in a continual manner has been
explored extensively in machine learning, seeded by the seminal
works of lifelong-learning \cite{thrun1995lifelong, thrun1995lifelong2,
  silver2013lifelong}, online-learning
\cite{fiat1998online,blum1998line,bottou1998online,bottou2004large} and
sequential linear gaussian models \cite{Roweis1999-ud,ghahramani2000online} such as the Kalman Filter
\cite{kalman1960new} and its non-linear counterpart, the Particle Filter \cite{del1996non}.
Lifelong learning bears some similarities to online learning in that
both learning paradigms observe data in a sequential manner. Online
learning differs from lifelong learning in that the central objective
of a typical online learner \cite{bottou1998online,bottou2004large} is to best solve/fit the current learning
problem, without preserving previous learning.
In contrast, lifelong learners seek to retain, and reuse, the learned behavior
acquired over past tasks, and aim to maximize performance across all tasks.
%
\note[Removed-18]{While lifelong learning has similarities to online learning, they
differ in their core objective. Typical online learners
\cite{bottou1998online,bottou2004large} aim to best model each
observed sequential task, while lifelong
learning attempts to model the entire space of tasks. }
Consider the example of forecasting click through rate: the objective of the online learner is
to evolve over time, such that it best represents \emph{current} user
preferences.
This contrasts lifelong learners which enforce a constraint between
tasks to ensure that previous learning is not lost.
\note[Removed-18]{We provide a more in depth comparison against state of the art online methods such as Streaming
Variational Bayes (SVB) \cite{DBLP:conf/nips/BroderickBWWJ13} and
incremental bayesian clustering methods \cite{katakis2008incremental,
  gomes2008incremental} in Appendix Section \ref{diff_streaming} for
the curious reader.}

Lifelong Learning \cite{thrun1995lifelong} was initially proposed in a
supervised learning framework for concept learning, where each
task seeks to learn a particular concept/class using binary classification.
The original framework used a task specific model, such as a K Nearest Neighbors (KNN)
\footnote{These models were known as memory based learning in \cite{thrun1995lifelong}.}, coupled with
a representation learning network
that used training data from all past learning tasks (support sets), to learn a common, global representation.
This supervised approach was later improved through the use of dynamic learning rates
\cite{silver1996parallel}, core-sets \cite{silver2015consolidation}
and multi-head classifiers \cite{fei2016learning}.
\note[Comment-18]{If time allows, describe briefly, a couple of sentences, how each one of them improved over the original? }
In parallel, lifelong learning was extended to independent multi-task
learning \cite{ruvolo2013ella,fei2016learning},
\note[Comment-18]{hhmmmm the Hall Daume work seems to be (from the title)
standard multi-task learning, does it really have a life-long flavor? The other
two seem to be about life-long. Standard multi-task requires (typically) seeing
everything together, so you should say, briefly, how this extension happened. }
reinforcement learning \cite{thrun1995lifelong2,tanaka1997approach,ring1997child},
topic modeling \cite{chen2014topic,wang2016mining} and semi-supervised language
learning \cite{mitchell2015never,mitchell2018never}. For a more
detailed review see \cite{chen2016lifelong}.

More recently, lifelong learning has seen a resurgence within the
framework of deep learning. As mentioned earlier, one of the central tenets of lifelong learning is that that the learner should
perform well over all observed tasks. Neural networks, and more
generally, models that learn using stochastic gradient descent
\cite{robbins1951stochastic}, typically cannot persist past task
learning without directly preserving past models or data. This problem
of catastrophic forgetting \cite{mccloskey1989catastrophic} is well
known in the neural network community and is the central obstacle that
needs to be resolved to build an effective neural lifelong learner.
%
Catastrophic forgetting is the
phenomenon where model parameters of a neural network
trained in a sequential manner become biased towards the
distribution of the latest observations, forgetting 
previously learned representations, over data no longer accessible for
training. In order to mitigate catastrophic forgetting
current research in lifelong learning 
employs four major strategies: 
\emph{transfer learning}, \emph{replay mechanisms}, \emph{parameter
  regularization} and \emph{distribution regularization}. 
In table \ref{strategy_table} we classify the different
lifelong learning methods that we will discuss in the following
paragraphs into these strategies.

\begin{table}[H]
  \begin{adjustbox}{width=\columnwidth}
    {\renewcommand{\arraystretch}{1.3}%
  \begin{tabular}{l|l|l|l|l|l|l|l|l|l|l|}
\cline{2-11}
                                                                                                                                                   &
                                                                                                                                                     \textbf{EWC
                                                                                                                                                     \cite{kirkpatrick2017overcoming}}
    & \textbf{VCL \cite{nguyen2018variational}}                   &
                                                                    \textbf{LwF \cite{li2016learning}}
    & \textbf{ALTM \cite{furlanello2016active}}                         & \textbf{PNN
                                              \cite{rusu2016progressive}}
    & \textbf{DGR \cite{shin2017continual, kamra2017deep}}                 & \textbf{DBMNN
                                     \cite{terekhov2015knowledge}} &
                                                                     \textbf{SI
                                                                     \cite{zenke2017continual}}
    & \textbf{VASE \cite{achille2018life}}                         & \textbf{LGM (us)}                     \\ \hline
\multicolumn{1}{|l|}{\textbf{Transfer learning}}                                                                                                   & \ding{55}                        & \ding{55}                        & \ding{55}                               & \ding{55}                               & \checkmark                          & \ding{55}                      & \checkmark         & \ding{55}                        & \ding{55}                               & \ding{55}                               \\ \hline
\rowcolor[HTML]{C0C0C0}
\multicolumn{1}{|l|}{\cellcolor[HTML]{C0C0C0}{\color[HTML]{000000} \textbf{Replay mechanisms}}}                                                    & \ding{55}                        & \checkmark                          & \ding{55}                               & \ding{55}                               & \ding{55}                        & {\color[HTML]{333333} \checkmark} & \ding{55}       & \ding{55}                        & \textbf{\checkmark}                        & \textbf{\checkmark}                        \\ \hline
\multicolumn{1}{|l|}{\textbf{\begin{tabular}[c]{@{}l@{}}Parameter \\ regularization\end{tabular}}}                                                 & \checkmark                          & \checkmark                          & \ding{55}                               & \ding{55}                               & \ding{55}                        & \ding{55}                      & \ding{55}       & \checkmark                          & \ding{55}                               & \ding{55}                               \\ \hline
\rowcolor[HTML]{C0C0C0}
\multicolumn{1}{|l|}{\cellcolor[HTML]{C0C0C0}{\color[HTML]{000000} \textbf{\begin{tabular}[c]{@{}l@{}}Functional \\ regularization\end{tabular}}}} & {\color[HTML]{333333} \ding{55}} & {\color[HTML]{333333} \ding{55}} & {\color[HTML]{333333} \textbf{\checkmark}} & {\color[HTML]{333333} \textbf{\checkmark}} & {\color[HTML]{333333} \ding{55}} & \ding{55}                      & \ding{55}       & {\color[HTML]{333333} \ding{55}} & {\color[HTML]{333333} \textbf{\checkmark}} & {\color[HTML]{333333} \textbf{\checkmark}} \\ \hline
\end{tabular}}
\end{adjustbox}
\caption{Catastropic interference mitigation strategies of state of the
  art models. Rows highlighted in gray represent desirable mitigation
  strategies.} \label{strategy_table}
\end{table}



\textbf{Transfer learning}: These approaches mitigate catastrophic
forgetting by freezing previous task models and relaying a
latent representation of the previous task to the current model. Research
in transfer learning for the mitigation of catastrophic forgetting include Progressive Neural Networks (PNN) \cite{rusu2016progressive} and Deep
Block-Modular Neural Networks (DBMNN)
\cite{terekhov2015knowledge} to name a few. 
These approaches allow the current model to adapt its parameters to
the (new) joint representation
\note[Comment-18]{what is "the (new) joint representation"? elaborate a bit
more on the two methods you reference. How can it be new and joint on the same time.
If it is new and joint past models need retraining.}
in an efficient manner and prevent forgetting through
the direct preservation of \emph{all previous task models}. 
\note[Rephrased-18]{ Due to 
 this, transfer learning mitigation approaches are unsuitable candidates for
long-lifetime (eg: \cite{mitchell2015never,mitchell2018never}) or
resource constrained learners such as those on embedded devices. 
}
Deploying such a transfer learning mechanism in a lifelong learning setting would
necessitate training a new model with every new task, considerably increasing
the memory footprint of the lifelong learner. In addition, since
transfer learning approaches freeze previous models, it negates the
possibility of improving previous task performance using
knowledge gathered from new tasks.

\textbf{Replay mechanisms}: The original formulation of lifelong
learning \cite{thrun1995lifelong} required the preservation of all
previous task data. This requirement was later relaxed in the form of
core-sets \cite{silver2002task,silver2015consolidation,nguyen2018variational},
which represent small weighted subsets of inputs that approximate the full dataset.
\note[rephrased-18]{Recently however there have been efforts to
use deep generative replay to prevent catastrophic forgetting in a
classification setting \cite{shin2017continual,kamra2017deep}.}
Recently, within the classification setting, there have been replay
approaches that try to lift the requirement of storing past training data
by relying on generative modeling \cite{shin2017continual,kamra2017deep};
we will call such methods Deep generative replay (DGR) methods.
DGR methods methods use a student-teacher network
architecture, where the teacher (generative) model augments the student
(classifier) model with synthetic samples from previous tasks. These synthetic task samples
are used in conjunction with real samples from the current task to learn a new joint model
across all tasks. While strongly motivated by biological rehearsal processes
\cite{skaggs1996replay,johnson2007neural,karlsson2009awake,schuck2019sequential},
these generative replay strategies fail to efficiently use previous
learning and simply re-learn each new joint task from scratch.



\textbf{Parameter regularization}:
Most work that mitigates catastrophic forgetting falls under the
umbrella of parameter regularization. There are two approaches within this
mitigation strategy: constraining the {parameters} of the new
task to be close to the previous task through a predefined metric,  and
enforcing task-specific parameter sparsity.
The two approaches are related as task-specific parameter sparsity can be perceived as
a refinement of the parameter constraining approach. 
Parameter constraining approaches typically share the same
model/parameters, but encourage new tasks from altering important
learned parameters from previous tasks.
Task specific parameter sparsity relaxes this, by enforcing
that each task use a different subset of parameters from a global
model, through the use of an attention mechanism.


Models such as Laplace Propagation \cite{eskin2004laplace}, Elastic Weight
Consolidation (EWC) \cite{kirkpatrick2017overcoming}, Synaptic
Intelligence (SI) \cite{zenke2017continual} and Variational Continual
Learning (VCL) \cite{nguyen2018variational} fall under the parameter constraining approach.
EWC for example,  
uses the Fisher Information matrix (FIM) to control the change of model parameters between two learning tasks.
Intuitively, important parameters should
not have their values changed, while non-important parameters are left
unconstrained. The FIM 
is used as a weighting in a quadratic parameter difference regularizer
under a Gaussianity assumption of the \emph{parameter} posterior. 
 However, this Gaussian \emph{parameter} posterior assumption has been demonstrated
 \cite{Neal1995-dx,blundell2015weight} to be sub-optimal for learned
 neural network parameters. 
 VCL improves upon EWC, by generalizing the local assumption of the FIM to a KL-Divergence
 between the (variational) \emph{parameter} posterior and prior. This
 generalization derives from the fact that the FIM can be cast as a
 KL divergence between the posterior and an epsilon perturbation of
 the same random variable \cite{jeffreys1946invariant}.
\note[Comment-18]{
If time allows:
1) eehmmm what is the local assumption of FIM? you did not discuss that in EWC.
2) What is a "non-local KL-divergence" elaborate and give also intuitions in addition
to the technical details.
}
VCL actually spans a number of different mitigation strategies as
it uses parameter regularization (described above)
, transfer learning (it keeps a separate head
network per task) and replay (it persists a core-set of true data per
task).


Models such as Hard Attention to the Task (HAT) \cite{serra2018overcoming} and the Variational Autoencoder with Shared
Embeddings (VASE) \cite{achille2018life} fall under the task-specific
parameter sparsity strategy.
This mitigation strategy enforces that different tasks use different
components of a single model, typically through the use of attention
vectors \cite{bahdanau2014neural} that are learned given supervised task labels.
Multiplying the attention vectors with the model outputs prevents gradient descent updates for different subsets
 of the model's parameters, allowing them to be used for future task learning. Task specific parameter sparsity allows a model to
hold-out a subset of its parameters for future learning and typically works well in
practice \cite{serra2018overcoming}, with its strongest disadvantage
being the requirement of supervised information.

\note[Rephrased-18]{
Most work that mitigates catastrophic forgetting falls under the umbrella of \emph{parameter
  regularization}. There exist two over-arching paradigms within this
mitigation strategy: enforcing task-specific \emph{parameter} sparsity and constraining the \emph{parameters} of the new
task to be close to the previous task through a predefined metric.

Models such as Hard Attention to the Task (HAT)
\cite{serra2018overcoming} and the Variational Autoencoder with Shared
Embeddings (VASE) \cite{achille2018life} use supervised information to learn a task specific attention vector \cite{bahdanau2014neural}
\footnote{Hard attention can be perceived as a one-hot vector with
  soft-attention \cite{bahdanau2014neural} being its continuous
  relaxation (typically through the softmax function).}. This attention vector is multiplied by
model outputs and prevents gradient descent updates of a subset of
parameters. Task specific parameter sparsity allows a model to
hold-out a subset of its parameters for future learning and typically works well in
practice \cite{serra2018overcoming}, with its strongest disadvantage
being the requirement of supervised information.


In contrast, models such as Elastic Weight Consolidation (EWC) \cite{kirkpatrick2017overcoming}, Synaptic
Intelligence (SI) \cite{zenke2017continual} and Variational Continual
Learning (VCL) \cite{nguyen2018variational} constrain model parameters
between tasks. EWC \cite{kirkpatrick2017overcoming} for example, uses the Fisher Information matrix (FIM)
to control the change of model parameters
between two tasks. Intuitively, important parameters should
not have their values changed, while non-important parameters are left
unconstrained. The FIM 
is used as a weighting in a quadratic parameter difference regularizer
under a Gaussianity assumption of the \emph{parameter} posterior. 
 This Gaussian \emph{parameter} posterior assumption has been demonstrated
 \cite{Neal1995-dx,blundell2015weight} to be sub-optimal for learned
 neural network parameters. 
 VCL improves upon EWC, by generalizing the local assumption of the FIM to a non-local KL-Divergence
 between the \emph{parameter} posterior and prior \cite{jeffreys1946invariant}.
However, VCL 
also adds a separate head network 
and a core-set of true data-samples \emph{per observed task},
preventing its scalability to long-lifetime or embedded devices.


}

\textbf{Functional regularization}:
Parameter regularization methods attempt to preserve the learned behavior
of the past models by controlling how the model parameters change
between tasks. However, the model parameters are only a proxy
for the way a model actually behaves. Models with very different
parameters can have exactly the same behavior with respect to
input-output relations (non-uniqueness \cite{williamson1995existence}).
Functional regularization concerns itself with preserving the actual
object of interest: the input-output relations. This strategy allows the model
to flexibly adapt its internal parameter representation between tasks,
while still preserving past learning.


Methods such as
distillation \cite{hinton2015distilling}, ALTM
\cite{furlanello2016active} and Learning Without Forgetting (LwF)
\cite{li2016learning} impose similarity constraints on the
classification outputs of models learned over different tasks. This
can be interpreted as functional regularization
by generalizing the constraining metric (or semi-metric) to be a
divergence on the output conditional distribution. 
In contrast to parameter regularization, no
assumptions are made on the parametric form of the parameter posterior distribution. This allows models to flexibly adapt their
internal representation as needed, making functional
  regularization a desirable mitigation strategy. One of the pitfalls
of current functional regularization approaches is that they
necessitate the preservation of \emph{all previously data}.


\vspace{-0.05in}
\subsection{Limitations of existing approaches.}
\vspace{-0.05in}


A simple solution to the problem of lifelong learning is to
 store all data and re-learn a new joint multi-task representation \cite{caruana1997multitask} at
 each newly observed task. Alternatively, it is possible to retain all
 previous model parameters and select the model that presents the best
 performance on new test task data. Existing solutions typically relax one
 these requirements. \cite{furlanello2016active,li2016learning,nguyen2018variational}
 relaxes the need for model persistence, but requires preservation of
 all data \cite{furlanello2016active,li2016learning}, or a growing
 core-set of data \cite{silver2002task,silver2015consolidation,nguyen2018variational}. Conversely,
 \cite{rusu2016progressive,terekhov2015knowledge,nguyen2018variational,zenke2017continual}
 relaxes the need to store data, but persists all previous models
 \cite{rusu2016progressive,terekhov2015knowledge} or a subset of model
 parameters \cite{nguyen2018variational,zenke2017continual}.
 \note[Alexandros]{Remove the two following sentences.}
 \note[Jason]{Roger.}

\note[Jason]{Roger. +Small fixes.}
\note[Add]{Unlike these approaches we draw inspiration from how humans learn over
time and seek to render unnecessary the storing of past training data
and models.}
Unlike these approaches, we draw inspiration from how humans learn over
time and remove the requirement of storing past training data and models.
Consider the human visual system; research has shown
\cite{curcio1990human,blackwell1946contrast} that the human eye is
capable of capturing 576 megapixels of content per image frame. If
stored naively on a traditional computer, this corresponds to
approximately 6.9 gigabytes of information per sample. Given that we
perceive trillions of frames over our lifetimes, it is infeasible to
store this information in its base, uncompressed
representation. Research in neuroscience has validated
\cite{wittrock1992generative,anderson2014human} that the associative
human mind, compresses, merges and reconstructs information content in
a dynamic way. Motivated by this, we believe that a lifelong learner should not
store past training data or models. Instead, it should retain a latent representation that is
common over all tasks and evolve it as more tasks are observed. 
\note[Jason]{Roger. +Small fixes. 'behavior' is unclear in this setting.}
\note[Alexandros]{Remove previous sentence}
\note[Add]{Motivated by this, as already mentioned, we believe that a lifelong learner should not
store past training data, nor it should models. Instead it should retain a latent representation that is
common over all tasks and evolve it as more tasks are seen.  The only constraint that we will impose
is that the learned representation should preserve past learned
behaviors.}

\vspace{-0.05in}
\subsection{Our solution at a high level.}
\vspace{-0.05in}

\note[Alexandros]{I would replace (or restructure) the whole subsection with the following.}
\note[Add]{
While most lifelong learning work focuses on supervised learning,
\cite{rusu2016progressive,furlanello2016active,li2016learning,terekhov2015knowledge,zenke2017continual,kirkpatrick2017overcoming},
we focus on the more challenging task of unsupervised learning and in particular deep generative modelling with latent variables.
At the core of our lifelong learning method we place a generative model, which we train by exploiting
replay and functional regularisation strategies.  Our generative model evolves as new tasks are
seen and allows us to generate, at will, data from any of the past distributions which we use for replay. As a
direct result it is not any more necessary to
store training data from any of the past distributions. Once a new task arrives we learn a new generative model,
which we call the student, over real data from the new task and replay data from the past tasks.
We generate the latter using the generative model learned over these past tasks, which we call the teacher. When
the student's training is completed it becomes on its turn the teacher, containing everything learned so far.
In order to preserve the past learned behaviors we make use of functional regularisation. We require that our
generative model behaves as it was trained over past tasks, in terms of input-output relations, when confronted
with (generated) data that correspond to these past tasks; we enforce this behaviour through the use of an
appropriate regulariser. Unlike EWC or VCL we make no parametric assumptions
and allow the generative model to use the available parameters as appropriate in order to be able to learn
all so far seen tasks in the best manner. The use of the generative replay mechanism coupled with our functional regularisation
renders unnecessary the preservation of the past models as well as the past data and brings in significant
performance gains in terms of sample complexity on the future tasks. Finally we should note that it is rather
straightforward to adapt our approach in the setting of supervised learning as in fact we have done in
\cite{DBLP:journals/corr/abs-1810-10612}.
}

While most research in lifelong learning focuses on supervised
learning \cite{rusu2016progressive,furlanello2016active,li2016learning,terekhov2015knowledge,zenke2017continual,kirkpatrick2017overcoming},
we focus on the more challenging task of deep unsupervised \emph{latent
variable} generative modeling. These models have wide ranging
applications such as
 clustering \cite{makhzani2015adversarial,jiang2017variational,nalisnick2017stick} and pre-training
\cite{larsen2016autoencoding,radford2015unsupervised}. 


Central to our lifelong learning method are a pair of generative
models, aptly named the teacher and student, which we train by exploiting the replay and functional regularization
strategies described above. After training a single generative model over the
first task, we use it is used as the teacher for a newly instantiated student model. The
student model receives data from the current task, as well as replayed
data from the teacher, which acts as a probabilistic storage container
of past tasks. In order to preserve previous learning, we make use of
functional regularization, which aids in preserving input-output
relations over past tasks. 

Unlike EWC or VCL, we make no assumptions on the form of the
parameter posterior and allow the generative models to use
 available parameters as appropriate, to best accommodate
current and past learning. The use of generative replay, coupled with
functional regularization, renders the preservation of the past models and past data
unnecessary. It also significantly improves the sample
complexity on future task learning, which we empirically demonstrate in our experiments. Finally we should note that it is
straightforward to adapt our approach to the supervised learning
setting, as we have done in \cite{DBLP:journals/corr/abs-1810-10612}.




%% file: background.tex
\section{Background} \label{background}
In this section we describe the main concepts that
we use throughout this work. We begin by describing the base-level generative modeling approach in Section
\ref{sec_generative_modeling}, followed by how it extends to the
lifelong setting in Section
\ref{sec_lifelong_generative_modeling}. Finally, in Section
\ref{sec_vae}, we describe the Variational Autoencoder over which we
instantiate our lifelong generative model.

\note[Rewrite]{In this section we describe the main concepts that
we use throughout this work. We begin by describing the base-level
generative modelling approach that we follow in Section
\ref{sec_generative_modeling} and how it extends to the lifelong
setting in Section \ref{sec_lifelong_generative_modeling}; in Section
\ref{sec_vae}, we describe the Variational Autoencoder model,
\cite{kingma2014}, which we have chosen to instantiate our generative model.}

\subsection{Latent Variable Generative Modeling}\label{sec_generative_modeling}

\begin{figure}[H]
  \begin{center}
    \includegraphics[width=0.4\linewidth]{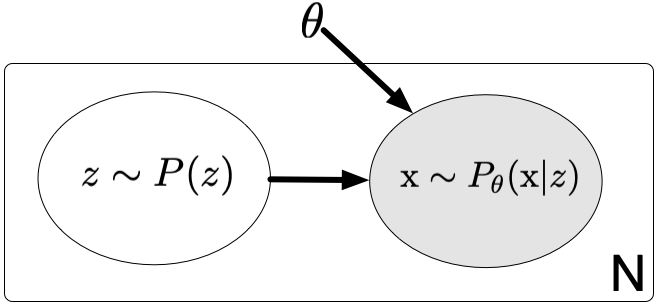}
  \end{center}
  \caption{Typical latent variable graphical model. Gray nodes
    represent observed variables while white nodes represent
    unobserved variables.}\label{gen_process}
  \vskip -0.1in
\end{figure}

We consider a scenario where we observe a dataset, $\bm{\mathcal{D}}$,
consisting of $N$ variates, $\bm{\mathcal{D}} =
\{\bm{\mathrm{x}}_j\}_{j=1}^N$, of a continuous or discrete variable
$\vect{x}$. We assume that the data is generated by a random
process involving a non-observed random variable, $\bm{z}$. The data
generation process involves first sampling $\bm{z}_j \sim P(\bm{z})$
and then producing a variate from the conditional, $\bm{\mathrm{x}}_j \sim
P_{\bm{\theta}}(\bm{\mathrm{x}} | \bm{z})$.
 We visualize this form of latent generative model in the graphical model in Figure \ref{gen_process}.

\vskip -0.1in

  Typically, latent variables models are solved through maximum likelihood
  estimation which can be formalized as:

\vskip -0.2in
\begin{align}
\max_{\bm{\theta}}\ \log P_{\theta}(\vect{x}) = \max_{\bm{\theta}}\ \log
  \int P_{\bm{\theta}}(\vect{x}| \bm{z}) P(\bm{z}) d \bm{z}  \
  = \max_{\bm{\theta}}\ \log
  \mathbb{E}_{\bm{z}}[P_{\bm{\theta}}(\bm{\mathrm{x}}|\bm{z})] \label{eqn_basic}
\end{align}

In many cases, the expectation from Equation \ref{eqn_basic} does not have a closed form solution (eg:
non-conjugate distributions) and quadrature is not computationally tractable due to large dimensional spaces
\cite{kingma2014,rezende2014stochastic} (eg: images). To overcome
these intractabilities we use a Variational Autoencoder (VAE), which
we summarize in Section \ref{sec_vae}. The VAE allows us to infer our latent variables and jointly estimate the parameters
of our model. However, before describing the VAE, it is important to understand how this
generative setting can be perceived in a lifelong learning scenario.

\subsection{Lifelong Generative Modeling}\label{sec_lifelong_generative_modeling}

Lifelong generative modeling extends the single-distribution
estimation task from Section \ref{sec_generative_modeling} to a set of
$i = \{1...L\}$ sequentially observed learning tasks.
The $i$-th learning task has variates that are realized from the
task specific conditional, $P_i(\bm{\mathrm{x}}) = P(\bm{\mathrm{x}} |
\bm{z}_d=i)$, where $\bm{z}_d$ acts as a categorical indicator variable of the
current task. We visualize  a simplified form of this in Figure
\ref{fig_problem_setup} below.

\begin{figure}[H]
\begin{center}
  \centerline{\includegraphics[width=\linewidth]{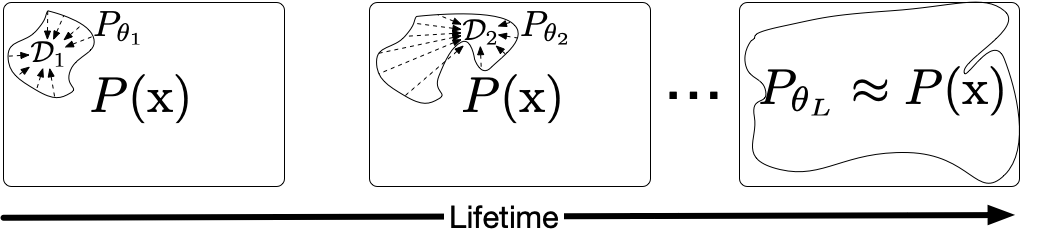}}
\caption{Simplified lifetime of a lifelong learner. Given a
  true (unknown) distribution,
  $P(\bm{\mathrm{x}}) = \int P(\bm{\mathrm{x}} | \bm{z}_d) P(\bm{z}_d) \delta
\bm{z}_d$, we observe partial information in the form of L sequential tasks,
$\{\bm{\mathcal{D}}_1 \mapsto \bm{\mathcal{D}}_2, ... \mapsto \bm{\mathcal{D}}_L\}$. Observing more tasks, reduces the uncertainty of
  the model until convergence, $P_{\theta_L}(\bm{\mathrm{x}}) \approx P(\bm{\mathrm{x}})$. }
\label{fig_problem_setup}
\end{center}
\vskip -0.2in
\end{figure}


Crucially, when observing task, $\bm{\mathcal{D}}_i$, the model has no access to any of
the previous task datasets, $\bm{\mathcal{D}}_{<i}$.
As the lifelong learner observes more tasks, $\{\bm{\mathcal{D}}_1
\mapsto \bm{\mathcal{D}}_2 \mapsto ... \mapsto
\bm{\mathcal{D}}_L\}$, it should improve its estimate of the true distribution,
$P_{\theta}(\bm{\mathrm{x}}) \approx P(\bm{\mathrm{x}}) = \int P(\bm{\mathrm{x}} | \bm{z}_d) P(\bm{z}_d) \delta
\bm{z}_d$, which is unknown at the start of training.

\note[Alexandros]{Remove what follows, untill the end of the paragraph. You have already said that,
plus you compare against archeology here, not fair.}
\note[Removed]{
The original formulation in \cite{thrun1995lifelong} models the
lifelong learning problem in two stages (albeit for a
  classification setting): learn a representation
using stored support sets, $g(\bm{\mathcal{D}}_{<i})$, and use
it to improve the $i$-th task estimate,
$P_{\theta_i}(g(\bm{\mathrm{x}}))$. This two
  step learning solution was interpreted by \cite{thrun1995lifelong} as a meta-learning approach
  \cite{kalousis2002algorithm, vilalta2002perspective}. In our
  formulation however, we require the learner to improve its estimate without the preservation of the support sets,
  $\bm{\mathcal{D}}_{<i}$, or the addition of a per-task model,
  $\{P_{\theta_i}(\bm{\mathrm{x}}) \approx P_{i}(\bm{\mathrm{x}})\}_{i=1}^L$.
}

\subsection{The Variational Autoencoder}\label{sec_vae}

As eluded to in Section \ref{sec_generative_modeling}, we would like
to infer the latent variables from the data. This can be realized as
an alternative form of Equation \ref{eqn_basic} in
the form of Bayes rule: $P_{\bm{\phi}}(\bm{z}| \bm{\mathrm{x}}) =
P_{\bm{\theta}}(\bm{\mathrm{x}}| \bm{z}) P(\bm{z}) /
P_{\bm{\theta}}(\vect{x})$, where $P_{\bm{\phi}}(\bm{z}|
\bm{\mathrm{x}})$ is referred to as the latent variable posterior and
$P_{\bm{\theta}}(\bm{\mathrm{x}}| \bm{z})$ as the likelihood.
One method of approximating the posterior, $P_{\bm{\phi}}(\bm{z}|
\bm{\mathrm{x}})$, is through MCMC sampling methods such as Gibbs
sampling \cite{gelfand1990sampling} or Hamiltonian MCMC
\cite{neal2011mcmc}. MCMC methods have the advantage that they provide
asymptotic guarantees \cite{DBLP:conf/uai/NeiswangerWX14} of
convergence to the true posterior, $P_{\phi}(\bm{z}|
\bm{\mathrm{x}})$. However in practice it is not possible to know when
convergence has been achieved. In addition, due to their Markovian
nature, they possess an inner loop, which makes it challenging to scale
for large scale datasets.

In contrast, Variational Inference (VI) \cite{jordan1999introduction}
 side-steps the intractability of the posterior by
approximating it with a tractable distribution family, $Q_{\bm{\phi}}(\bm{z}|\bm{\mathrm{x}})$.
VI rephrases the objective of determining the posterior as an
optimization problem by minimizing the KL divergence
between the known distributional family, $Q_{\bm{\phi}}(\bm{z}|\bm{\mathrm{x}})$, and the unknown true
posterior, $P_{\bm{\theta}}(\bm{z}|\bm{\mathrm{x}})$. 
 Applying VI to the intractable integral from Equation \ref{eqn_basic} results in the evidence lower
 bound (ELBO) or variational free energy, which can easily be derived from first principles:
\begin{align}
\log P_{\bm{\theta}}(\bm{\mathrm{x}}) &= \log \int P_{\bm{\theta}}(\bm{\mathrm{x}}|\bm{z})P(\bm{z}) d \bm{z} \\
&= \log \int
  \frac{Q_{\bm{\phi}}(\bm{z}|\bm{\mathrm{x}})}{Q_{\bm{\phi}}(\bm{z}|\bm{\mathrm{x}})}
  P_{\bm{\theta}}(\bm{\mathrm{x}}|\bm{z})P(\bm{z}) d \bm{z} \label{elbo_eqn_m_1}\\
  &\geq \underbrace{\mathbb{E}_{Q}[\log P_{\bm{\theta}}(\bm{\mathrm{x}} |
  \bm{z})]-\mathbb{D}_{KL} [Q_{\bm{\phi}}(\bm{z}|\bm{\mathrm{x}}) ||
  P(\bm{z})]}_{\text{ELBO}} \label{elbo_eqn}
\end{align}
where we used Jensen's inequality to transition from
Equation \ref{elbo_eqn_m_1} to Equation \ref{elbo_eqn}.
The objective introduced in Equation \ref{elbo_eqn} induces the
graphical model shown below in Figure \ref{vae_graphical_model}.
%
\begin{figure}[H]
  \begin{center}
    \includegraphics[width=0.75\linewidth]{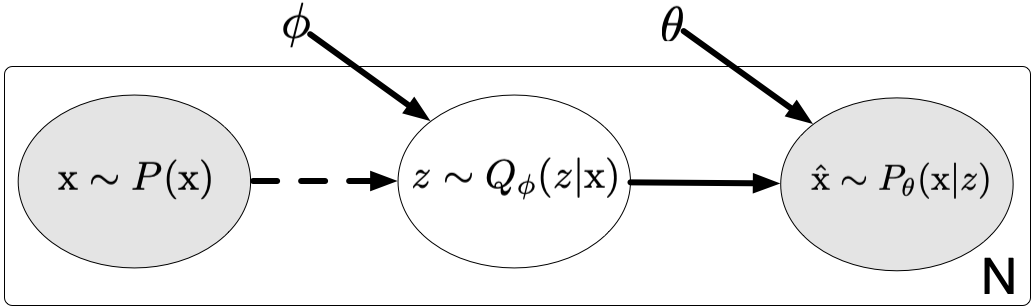}
  \end{center}
  \vskip -0.1in
  \caption{Standard VAE graphical model. Gray nodes
    represent observed variables while white nodes represent
    unobserved variables; dashed lines represent inferred variables.}\label{vae_graphical_model}
\end{figure}
\vskip -0.1in

VAEs typically use deep neural networks to model the approximate
inference network, $Q_{\bm{\phi}}(\bm{z}|\bm{\mathrm{x}})$ and
conditional, $P_{\bm{\theta}}(\bm{\mathrm{x}} | \bm{z})$, which are
also known as the encoder and decoder networks (respectively). To
optimize for the parameters of these networks, VAEs maximize the ELBO (Equation
\ref{elbo_eqn}) using Stochastic Gradient Descent \cite{robbins1951stochastic}. By sharing the variational parameters of the encoder, $\bm{\phi}$, across the
data points (\emph{amortized inference} \cite{gershman2014amortized}),
variational autoencoders avoid per-data inner loops typically
needed by MCMC approaches.
\note[Alexandros]{Now this is the appropriate place to speak about optimization,
since you are discussing a specific method.

By the way this discussion just bellow concerns only the expectation term over Q in
the ELBO. Make that clear right away. Say something like: Optimizing the ELBO objective  requires computing the
gradient over an expectation in which we sample over the $ Q_{\bm{\phi}}(\bm{z}|\bm{\mathrm{x}}$. the sampling
part does not allow the gradients to flow back to the encoder network. The gradient of the KL term does not have this
issue since it has a closed form solution in the case of isotropic gaussian distributions. The standard
way to address the former is through the use of path-wise...}

Optimizing the ELBO in Equation \ref{elbo_eqn} requires computing the
gradient of an expectation over the approximate posterior, $Q_{\bm{\phi}}(\bm{z} | \vect{x})$.
This typically takes place through the use of the path-wise estimator \cite{rezende2014stochastic,kingma2014} (originally called ``push-out''
\cite{rubinstein1992sensitivity}).
\note[Alexandros]{And this is what I call usefull information which I can go and read (though might be a bit too technical
for the lifelong setting). But still usefull and in context.}
The path-wise reparameterizer uses the Law of the Unconscious Statistician
(LOTUS) \cite{grimmett2001probability}, which enables us to compute the expectation of a function of a
random variable (without knowing its distribution) if we know its
corresponding \emph{sampling path} and \emph{base distribution} \cite{DBLP:journals/corr/abs-1906-10652}. For the typical
isotropic gaussian approximate posterior,
$Q_{\bm{\phi}}(\bm{z}|\bm{\mathrm{x}})$, used in standard VAEs this
can be aptly summarized by:
\begin{align}
  \bm{z} \sim Q_{\bm{\phi}}(\bm{z}|\bm{\mathrm{x}}) &\logeq
  \bm{\mu}_{\phi}(\bm{\mathrm{x}}) + \bm{\sigma}_{\bm{\phi}}(\bm{\mathrm{x}}) \epsilon,\ \ \epsilon \sim \mathcal{N}(0, 1) \label{reparam_eqn}
  \\
  \nabla_{\bm{\phi}} \mathbb{E}_{Q_{\bm{\phi}}(\bm{z} |
  \bm{\mathrm{x}})} [\log P_{\bm{\theta}}(\bm{\mathrm{x}} | \bm{z})]
  &\logeq \mathbb{E}_{\underbrace{\mathcal{N}(\epsilon | 0,
    1)}_{\text{base distribution}}} [\nabla_{\bm{\phi}}
  \log P_{\bm{\theta}}(\bm{\mathrm{x}} |
    \underbrace{\bm{\mu}_{\phi}(\bm{\mathrm{x}}) +
    \bm{\sigma}_{\bm{\phi}}(\bm{\mathrm{x}}) \epsilon}_{\text{sampling
    path}})] \label{path_eqn}
\end{align}
where Equation \ref{reparam_eqn} defines the sampling
procedure of our latent variable through the location-scale
transformation and Equation \ref{path_eqn} defines the path-wise
Monte Carlo gradient estimator applied on the decoder (first term in Equation
\ref{elbo_eqn}). This Monte Carlo estimator enables differentiating
through the sampling process of the distribution $Q_{\bm{\phi}}(\bm{z}
| \bm{\mathrm{x}})$. Note that computing the gradient of the second term in Equation \ref{elbo_eqn}, $\nabla_{\bm{\phi}} \mathbb{D}_{KL} [Q_{\bm{\phi}}(\bm{z}|\bm{\mathrm{x}}) ||
  P(\bm{z})]$, is possible through a closed form analytical solution
  for the case of isotropic gaussian distributions.

\note[Alexandros]{I wonder whether this VAE motivation should not have
come earlier in the subsection, before describing how in fact VAE operates.
I would also add here something that says why the latent variables per se
are useful in a life-long setting; as we said they allow for a finer control
of where to generate data from, i.e. from which task.}
While it is possible to extend any latent variable generative model to
the lifelong setting, we choose to build our lifelong generative models
using variational autoencoders (VAEs) \cite{kingma2014} as they
provide a mechanism for stable training; this contrasts other state of
the art unsupervised models such as Generative
Adversarial Networks (GANs)
\cite{goodfellow2014generative,kim2018disentangling}. Furthermore, latent-variable posterior approximations are a requirement in many learning scenarios such as clustering
\cite{quintana2003bayesian}, compression \cite{perlmutter1996bayes}
and unsupervised representation learning \cite{fe2003bayesian}. Finally, GANs can suffer from low sample diversity
\cite{dupont2018learning} which can lead to compounding errors in a
lifelong generative setting.

%% file: model.tex
\section{Lifelong Learning Model}\label{model_defn}

\begin{figure}[H]
  \minipage{0.60\textwidth}
  \includegraphics[width=\linewidth]{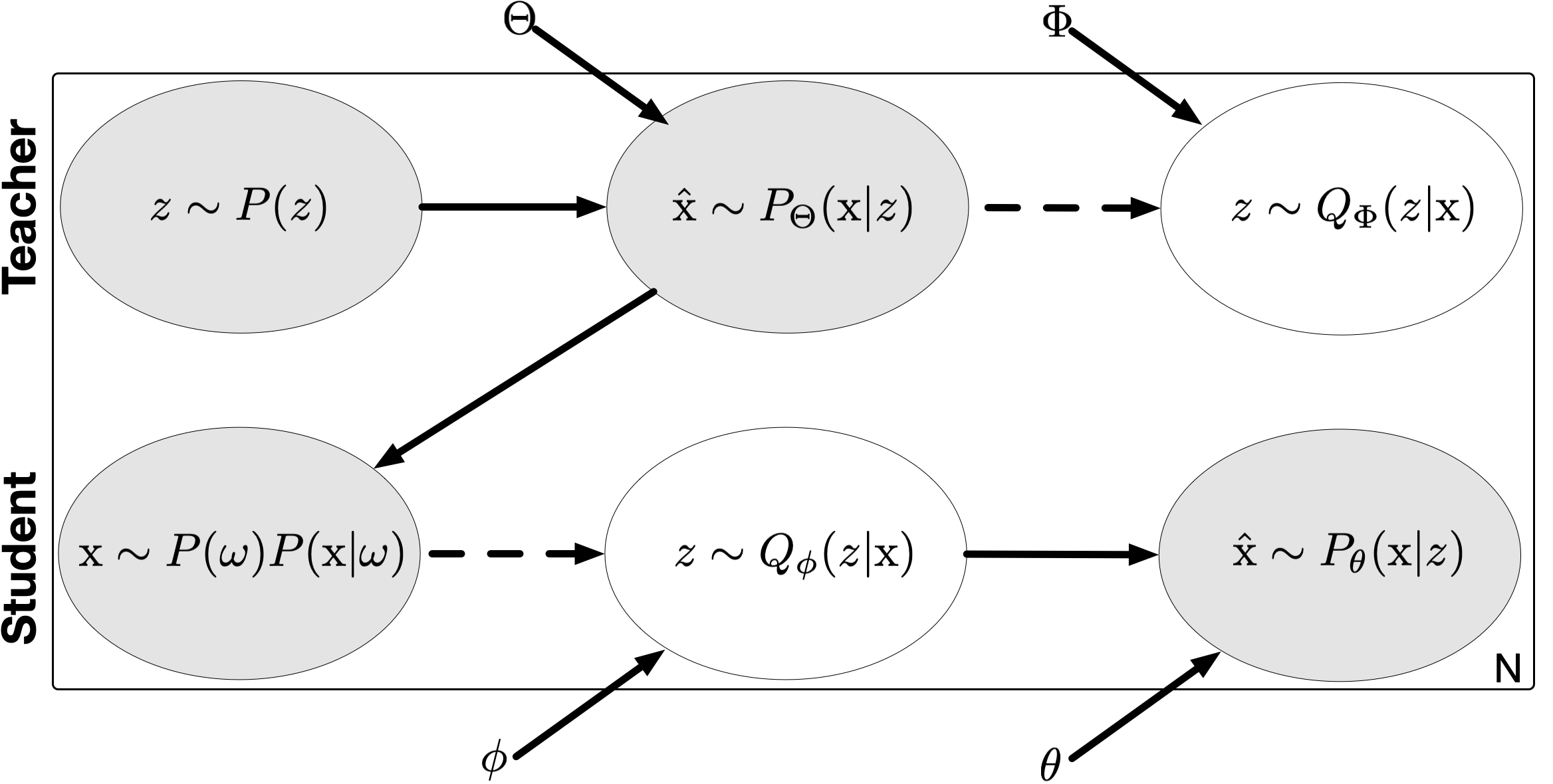}
  \endminipage\hfill
  \minipage{0.40\textwidth}
  \begin{algorithm}[H]
   \caption{Data Flow}
   \begin{algorithmic}
     \small
     \STATE {\bfseries Teacher:}
         \STATE {Sample Prior:} $\bm{z}_j \sim P(\bm{z})$
         \STATE {Decode:} $\hat{\bm{\mathrm{x}}}_j \sim
         P_{\bm{\Theta}}(\bm{\mathrm{x}} | \bm{z})$
         \note[Removed-16]{\STATE {Encode:} $\bm{z}_j \sim
           Q_{\bm{\Phi}}(\bm{z}|\bm{\mathrm{x}})$}
         \STATE {}
         \STATE {\bfseries Student:}
         \STATE {Sample :} $\bm{\mathrm{x}}_j \sim P(\bm{\omega})P(\bm{\mathrm{x}} | \bm{\omega})$
         \STATE {Encode :} $\bm{z}_j \sim Q_{\bm{\phi}}(\bm{z}|\bm{\mathrm{x}})$
         \STATE {Decode:} $\hat{\bm{\mathrm{x}}}_j \sim P_{\bm{\theta}}(\bm{\mathrm{x}} | \bm{z})$
\end{algorithmic}
\end{algorithm}
\endminipage\hfill
\caption{Student training procedure. \emph{Left}: graphical model for
  student-teacher model. Data generated from the teacher model (top
  row) is used to augment the current training data observed by the
  student model (bottom row). A posterior regularizer is also applied
  between $Q_{\bm{\phi}}(\bm{z}|\bm{\mathrm{x}})$ and $Q_{\bm{\Phi}}(\bm{z}|\bm{\mathrm{x}})$ to enable functional regularization (not shown, but discussed
  in detail in Section \ref{regularizer}).
      \emph{Right}: data flow algorithm.} \label{online-model}
  \end{figure}

fMRI studies of the rodent
\cite{skaggs1996replay,johnson2007neural,karlsson2009awake} and human
\cite{schuck2019sequential} brains have shown that previously
experienced sequences of events are replayed in the hippocampus during
rest. These replays are necessary for better planning
\cite{johnson2007neural} and memory consolidation
\cite{carr2011hippocampal}. We take inspiration from the memory
consolidation of biological learners and introduce our model of
Lifelong Generative Modeling (LGM). We visualize the LGM
student-teacher architecture in Figure \ref{online-model}.

The student and the teacher are both instantiations of the same
base-level generative model, but have different roles throughout the learning
process. The teacher's role is to act as a probabilistic knowledge
store of previously learned distributions, which it transfers to the
student in the form of replay and functional regularization.
 The student's role is to learn the distribution
over the new task, while accommodating the learned representation of
the teacher over old tasks. In the following sections we provide
detailed descriptions of the student-teacher architecture, as well as
the base-level generative model that each of them use. The base-level
model uses a variant of VAEs, which we tailor for lifelong
learning and is learned by maximizing a variant of the standard VAE
ELBO from Equation \ref{elbo_eqn} ;  we describe this objective at end of this section.


\subsection{Student-teacher Architecture}

The top row of Figure \ref{online-model} represents the teacher model.
At any given time, the teacher contains a summary of all previous
distributions within the learned parameters, $\bm{\Phi}$, of the encoder
$Q_{\bm{\Phi}}(\bm{z}|\bm{\mathrm{x}})$, and the learned parameters,
$\bm{\Theta}$, of the decoder $P_{\bm{\Theta}}(\bm{\mathrm{x}} | \bm{z})$.
We use the teacher to generate synthetic variates,
$\hat{\bm{\mathrm{x}}}_j$, from these past distributions by decoding
variates from the prior, $\bm{z}_j \sim P(\bm{z}) \mapsto
P_{\bm{\Theta}}(\bm{\mathrm{x}} | \bm{z} = \bm{z}_j)$.
We pass the generated (synthetic) variates, $\hat{\bm{\mathrm{x}}}_j$, to the student model as a
form of knowledge transfer about the past distributions. Information
transfer in this manner is known as \emph{generative replay} and our
work is the first to explore it in a VAE setting.

The bottom row of Figure \ref{online-model} represents the
student. The student is responsible for updating the parameters, $\bm{\phi}$, of its encoder,
$Q_{\bm{\phi}}(\bm{z}|\bm{\mathrm{x}})$, and $\bm{\theta}$, of its decoder
$P_{\bm{\theta}}(\bm{\mathrm{x}} | \bm{z})$. Importantly, the student
receives data from both the currently observed task, as well as
synthetic data generated by the teacher. This can be formalized as $\bm{\mathrm{x}}_j \sim
P(\bm{\omega})P(\bm{\mathrm{x}} | \bm{\omega}), \, \bm{\omega} \sim
\text{Ber}(\pi)$, as shown in Equation \ref{xsampling}:

\vskip -0.2in
\begin{align}
P(\bm{\omega})P(\bm{\mathrm{x}} | \bm{\omega}) = \begin{cases}
      P_{\bm{\Theta}}(\bm{\mathrm{x}}|\bm{z}) & \bm{\omega}=0 \\
      P_i(\bm{\mathrm{x}}) & \bm{\omega}=1 \label{xsampling}
\end{cases}
\end{align}
The mean, $\pi$, of the Bernoulli distribution, controls
the sampling proportion of the previously learned distributions to the
current one and is set based on the number of assimilated
distributions. 
Thus, given $i$ observed distributions: $\pi = \frac{1}{i+1}$.
This ensures that the samples observed by the student are
representative of both the current and past distributions.
Note that this does not correspond to varying sample sizes in datasets, but merely our
assumption to model each distribution with equivalent weighting.


Once a new task is observed, the old teacher is dropped, the student model is frozen and becomes
the new teacher ($\bm{\phi} \to \bm{\Phi}, \bm{\theta} \to
\bm{\Theta}$). A new student is then instantiated with the latest weights $\bm{\phi}$ and $\bm{\theta}$
from the previous student (the new teacher). Due to the cyclic nature
of this process, no new models are added. This contrasts many existing
state of the art deep lifelong learning methods which add an entire
new model or head-network per task (eg: \cite{nguyen2018variational,rusu2016progressive,terekhov2015knowledge}).


A crucial aspect in the lifelong learning process is to ensure that
previous learning is successfully exploited to bias current learning
\cite{thrun1995lifelong}.
While the replay mechanism that we
put in place ensures that the student will observe data from all tasks, it does not ensure
that previous knowledge from the teacher is efficiently exploited to
improve current student learning. The student model will re-learn (from scratch) a
completely new representation, which might be different than the teacher. In order to
successfully transfer knowledge between both VAE models, we rely on
\emph{functional regularization}, which we enforce through a Bayesian
update regularizer of the posteriors of both models. Intuitively, we
would like the student model's latent outputs, $\bm{z}_j \sim
Q_{\bm{\phi}}(\bm{z}| \vect{x})$ to be similar to latent outputs of
teacher model, $\bm{z}_j \sim Q_{\bm{\Phi}}(\bm{z}| \vect{x})$, over
synthetic variates generated by the teacher, $\vect{x}_j \sim
P(\bm{\omega})P(\bm{\mathrm{x}} | \bm{\omega}=0) = P_{\bm{\Theta}}(\vect{x} | \bm{z})$.
In the following section, we describe the exact functional form of
this regularizer and demonstrate how it can be perceived as a natural
extension of the VAE learning objective to a sequential setting.


\subsubsection{Knowledge Transfer Via Bayesian
  Update.}\label{regularizer}
\note[Jason]{Added it to the base of Section 4.0}
\note[Alexandros-16]{Ahahaha... before introducing the regulariser we need to
  say how actually the student is trained... Check.}

While both the student and teacher are instantiations of VAE variants,
tailored for the particularities of the lifelong setting, for the
purpose of this exposition we use the standard VAE formulation.
Our objective is to learn the set of parameters $[\bm{\phi},
\bm{\theta}]$ of the student, such that it can generate
variates from the complete distribution, $P(\bm{\mathrm{x}})$, described in Section
\ref{sec_lifelong_generative_modeling}. Subsuming the definition of
the augmented input data, $\vect{x} \sim P(\bm{\omega})P(\bm{\mathrm{x}} | \bm{\omega})$, from Equation \ref{xsampling}, we can define
the student ELBO as:
\begin{align}
\label{studentELBO}
  {\mathcal{L}}_{\bm{\theta}, \bm{\phi}}(\vect x) =
   &\ \mathbb{E}_{Q_{\bm{\phi}}(\bm{z}|\vect x)} \bigg[ \log P_{\bm{\theta}}(\vect{x}| \bm{z})\bigg] -
                                                         \text{KL}
     [Q_{\bm{\phi}}(\bm{z}|\vect x) || P(\bm{z})],  \\
& \bm{\mathrm{x}} \sim P(\bm{\omega})P(\bm{\mathrm{x}} | \bm{\omega}), \, \bm{\omega} \sim \text{Ber}(\pi). \nonumber
\end{align}
Rather than naively shrinking the full posterior to the prior via the
KL divergence in Equation \ref{studentELBO}, we rely on one of the core
tenets of the Bayesian paradigm which states that we can always update
our posterior when given new information (“yesterday’s posterior is
today’s prior”) \cite{mcinerney2015population}. Given this tenet, we
introduce our posterior regularizer \footnote{While it is also possible to apply a similar regularizer to the
reconstruction term, i.e: $KL[P_{\bm{\theta}}(\bm{\mathrm{x}} |
\bm{z})\ || \ P_{\bm{\Theta}}(\bm{\mathrm{x}} | \bm{z})]$,
we observed that doing so hurts performance (Appendix \ref{ll_regularizer}).}:
\begin{align}
  KL[ Q_{\bm{\phi}}(\bm{z} |    {\bm{\mathrm{x}}}     ) || Q_{\bm{\Phi}}(\bm{z}
  |    {\bm{\mathrm{x}}}    ) ],\ \vect x \sim P(\bm{\mathrm{x}} | \bm{\omega=0}) \label{posterior_reg_eqn-v1}
\end{align}
which distills the teacher's learnt representation into the student \emph{over the generated data only}. 
Combining Equations \ref{studentELBO} and \ref{posterior_reg_eqn-v1},
yields the objective that we can use to train the student and is
described below in Equation \ref{student_elbo_bayes_eqn}:
\begin{align}
  \label{student_elbo_bayes_eqn}
  {\mathcal{L}}_{\bm{\theta}, \bm{\phi}}(\vect x) =
   & \ \mathbb{E}_{Q_{\bm{\phi}}(\bm{z}|\vect x)} \bigg[ \log
     P_{\bm{\theta}}(\vect{x}| \bm{z}) \bigg] -
                                                         \text{KL} [Q_{\bm{\phi}}(\bm{z}|\vect{x}) || P(\bm{z})] \\ \nonumber
   & +  (1-\omega)\text{KL}[ Q_{\bm{\phi}}(\bm{z} |    {\bm{\mathrm{x}}}     ) || Q_{\bm{\Phi}}(\bm{z}
  |    {\bm{\mathrm{x}}}    ) ] ,  \\
& \bm{\mathrm{x}} \sim P(\bm{\omega})P(\bm{\mathrm{x}} | \bm{\omega}), \, \bm{\omega} \sim \text{Ber}(\pi) \nonumber
\end{align}
Note that this is not the final objective, due to the fact that we
have yet to present the VAE variant tailored to the particularities of the lifelong setting.
We will now show how the posterior regularizer can be perceived as a natural
extension of the VAE learning objective, through the lens of a Bayesian update of the student posterior.

\note[Alexandros-16]{In the lemma you can simply use $\vect x$ instead of $\hat{\vect x}$,
after the lemma I just keep $\vect x$ and $\hat{\vect x}$, i.e. drom from the latter the $_{<i}$ index}

\begin{lemma}
  For random variables $\bm{\mathrm{x}}$ and $\bm{z}$ with conditionals
  $Q_{\bm{\Phi}}(\bm{z} | \bm{\mathrm{x}})$ and
  $Q_{\bm{\phi}}(\bm{z} | \bm{\mathrm{x}})$, both distributed
  as a categorical or gaussian  
  and parameterized by
  $\bm{\Phi}$ and $\bm{\phi}$ respectively
      , the KL
  divergence between the distributions is: 
  \begin{align}
    \begin{split}
    KL[ Q_{\bm{\phi}}(\bm{z} | \bm{\mathrm{x}}) ||
    Q_{\bm{\Phi}}(\bm{z} | \bm{\mathrm{x}}) ] = KL[
    Q_{\hat{\bm{\phi}}}(\bm{z} | \bm{\mathrm{x}}) || P(\bm{z})]
    + C(\bm{\Phi})
    \end{split}
  \end{align}

where $\hat{\bm{\phi}} = f(\bm{\phi}, \bm{\Phi})$ depends on the
parametric form of Q, and C is only a function of the parameters, $\bm{\Phi}$.
\end{lemma}\label{lemma}

We prove Lemma 1 for the relevant distributions (under some mild
assumptions) in Appendix \ref{continuous_regularizer}. Using Lemma 1
allows us to rewrite Equation \ref{student_elbo_bayes_eqn} as shown
below in Equation \ref{student_elbo_bayes_eqn_v2}:
\begin{align}
  \label{student_elbo_bayes_eqn_v2}
  {\mathcal{L}}_{\bm{\theta}, \bm{\phi}}(\vect x) =
   & \ \mathbb{E}_{Q_{\bm{\phi}}(\bm{z}|\vect x)} \bigg[ \log
     P_{\bm{\theta}}(\vect{x}| \bm{z}) \bigg] -
                                                         \text{KL} [Q_{\bm{\phi}}(\bm{z}|\vect{x}) || P(\bm{z})] \\ \nonumber
   & +  (1-\omega)\bigg[\text{KL}[ Q_{\hat{\bm{\phi}}}(\bm{z} |
     {\bm{\mathrm{x}}}     ) || P(\bm{z}) ] + C(\bm{\Phi})\bigg] ,  \\
& \bm{\mathrm{x}} \sim P(\bm{\omega})P(\bm{\mathrm{x}} | \bm{\omega}), \, \bm{\omega} \sim \text{Ber}(\pi) \nonumber
\end{align}
This rewrite makes it easy to see that our posterior regularizer
from Equation \ref{student_elbo_bayes_eqn} is a standard VAE ELBO
(Equation \ref{elbo_eqn}) under a reparameterization of the student
parameters, $\hat{\bm{\phi}} = f(\bm{\phi}, \bm{\Phi})$. Note that $C(\bm{\Phi})$ is constant with respect to the
student parameters, $\bm{\phi}$, and thus not used during
optimization. While the change seems minor, it omits the introduction of
$f(\bm{\phi}, \bm{\Phi})$ which allows for a transfer of information
between models. In practice, we simply analytically evaluate $KL[
Q_{\bm{\phi}}(\bm{z} | \bm{\mathrm{x}})\ ||
    Q_{\bm{\Phi}}(\bm{z} | \bm{\mathrm{x}}) ]$, the KL
divergence between the teacher and the student posteriors,
instead of deriving the functional form of $f(\bm{\phi}, \bm{\Phi})$
for each different distribution pair.
We present Equation \ref{student_elbo_bayes_eqn_v2} simply as a means
to provide a more intuitive understanding of our functional regularizer.

\subsection{Base-level generative model.}\label{base_gen_sec}

\begin{figure*}[ht]
  \begin{center}
    \vskip -0.3in
    \includegraphics[width=\linewidth]{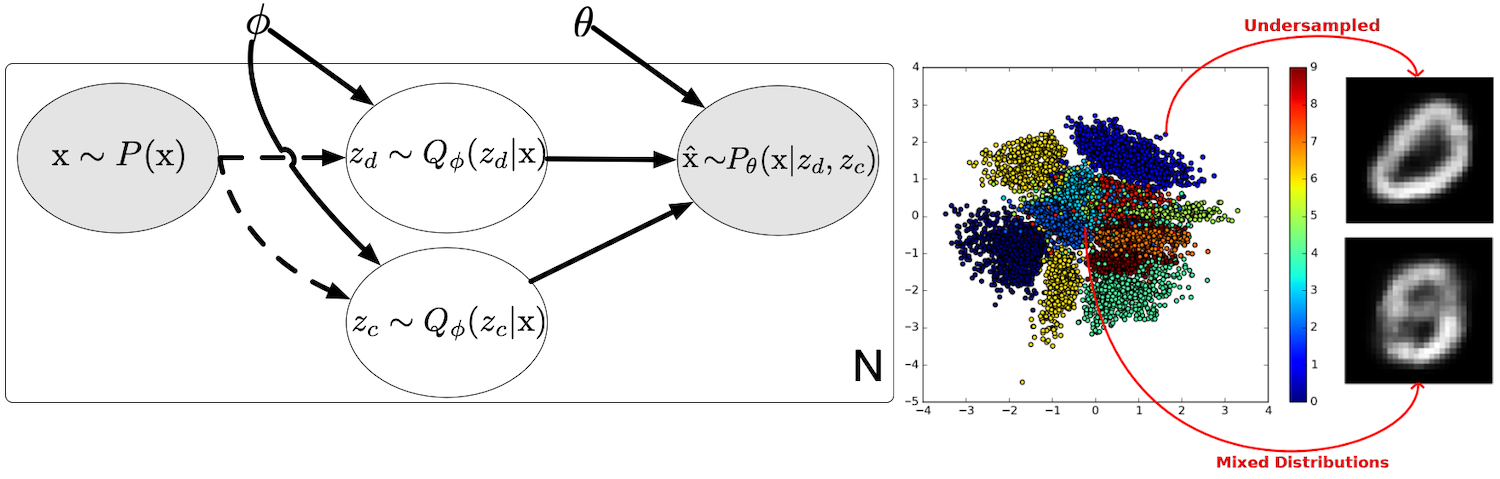}
  \end{center}
  \vskip -0.2in
  \caption{\emph{Left}: Graphical model for VAE with independent
    discrete and continuous posterior, $Q_{\bm{\phi}}(\bm{z}_c,
    \bm{z}_d | \vect{x}) = Q_{\bm{\phi}}(\bm{z}_c |
    \vect{x})Q_{\bm{\phi}}(\bm{z}_d | \vect{x})$. \emph{Right}: Two
    dimensional \emph{test} variates, $\bm{z}_j \sim Q_{\bm{\phi}}(\bm{z} |
    \vect{x})$, $\bm{z}_j \in \mathbb{R}^2$, of a vanilla VAE trained on MNIST. We depict the
  two generative shortcomings visually: 1) mixing of distributions which causes
  aliasing in a lifelong setting and 2) undersampling of distributions in
  a standard isotropic-gaussian VAE
  posterior.} \label{vae-limitations}
\vskip -0.2in
\end{figure*}

\note[Alexandros-16]{Check the following. It looks to me that at least for the moment it is better here.}
While it is theoretically possible to use the vanilla VAE from Section \ref{sec_vae} for
the teacher and student models, doing so brings to light a number of limitations that render
it problematic for use in the context of lifelong
learning (visualized in Figure
\ref{vae-limitations}-\emph{Right}). Specifically, using a standard
VAE decoder, $P_{\bm{\theta}}(\vect{x}|\bm{z})$, to generate synthetic
replay data for the student is problematic
due to two reasons:
\begin{enumerate}
  \item \textbf{Mixed Distributions}: Sampling the continuous standard normal
    prior, $\mathcal{N}(0, 1)$, can select a point in latent space that is in between two
separate distributions, causing generation of unrealistic synthetic data and eventually
leading to loss of previously learnt distributions.
\item \textbf{Undersampling}: Data points mapped to the
  isotropic-gaussian posterior that are further away from the prior mean will be sampled less frequently,
resulting in an undersampling of some of the constituent distributions.
\end{enumerate}

To address these sampling limitations we decompose the latent
variable, $\bm{z}$, into an independent continuous, $\bm{z}_c \sim
Q_{\bm{\phi}}(\bm{z}_c | \vect{x})$, and a discrete
component, $\bm{z}_d \sim Q_{\bm{\phi}}(\bm{z}_d| \vect{x})$, as shown
in Equation \ref{decompose_eqn} and visually in Figure \ref{vae-limitations}-\emph{Left}:
\begin{align}
\label{decompose_eqn}
Q_{\bm{\phi}}(\bm{z}_c,
    \bm{z}_d | \vect{x}) = Q_{\bm{\phi}}(\bm{z}_c |
    \vect{x})Q_{\bm{\phi}}(\bm{z}_d | \vect{x}).
\end{align}

\noindent The objective of the discrete component is to summarize the discriminative
information of the individual generative distributions. The continuous component on the other hand,
caters for the remaining sample variability (a nuisance variable
\cite{louizos2015variational}). Given that the discrete component can
accurately summarize the discriminative information, we can then explicitly sample from any
of the past distributions, allowing us to balance the student model's
synthetic inputs with samples from all of the previous learned distributions.
 We describe this beneficial generative sampling property in more detail in Section \ref{sampling}.

\note[Alexandros-16]{Check}
Naively introducing the discrete component, $\bm{z}_d$, does not guarantee
that the decoder will use it to represent the most discriminative
aspects of the modeled distribution. In preliminary experiments, we observed that that the decoder
typically learns to ignore the discrete component and simply relies on
the continuous variable, $\bm{z}_c$. This is similar to the posterior
collapse phenomenon which has received a lot of recent interest within
the VAE community \cite{razavi2019preventing,goyal2017z}. Posterior
collapse occurs when training a VAE  with a
powerful decoder model such as a PixelCNN++
\cite{tomczak2018vae} or RNN
\cite{chung2015recurrent,goyal2017z}. The output of the decoder,
$\vect{x}_j \sim P_{\bm{\theta}}(\vect{x} | \bm{z})$ can become almost
independent of the posterior sample, $\bm{z}_j \sim
Q_{\bm{\phi}}(\bm{z} | \vect{x})$, but is still able to reconstruct
the original sample by relying on its auto-regressive property \cite{goyal2017z}.
In Section \ref{information_regularizer}, we introduce a mutual
information regulariser which ensures that the discrete component of
the latent variable is not ignored.

\subsubsection{Controlled Generations.}\label{sampling}


\vskip -0.3in
\begin{figure}[H]
  \begin{minipage}{0.62\textwidth}
    \centering
    \begin{adjustbox}{width=\columnwidth}
      \begin{tabular}{l|l|l|}
        \cline{2-3}
                                       Desired Task Conditional & \textbf{$\bm{z}_c$} & \textbf{$\bm{z}_d$} \\ \hline
        \multicolumn{1}{|l|}{\textbf{$P_1(\vect{x}) = P(\bm{\mathrm{x}} | \bm{z}_d=1)$}} & $\sim \mathcal{N}(0,1)$        & {[}0, 0, 1{]} \\ \hline
        \multicolumn{1}{|l|}{\textbf{$P_2(\vect{x}) = P(\bm{\mathrm{x}} | \bm{z}_d=2)$}} & $\sim \mathcal{N}(0,1)$        & {[}0, 1, 0{]} \\ \hline
        \multicolumn{1}{|l|}{\textbf{$P_3(\vect{x}) = P(\bm{\mathrm{x}} | \bm{z}_d=3)$}} & $\sim \mathcal{N}(0,1)$        & {[}1, 0, 0{]} \\ \hline
      \end{tabular}\label{table_gen_example}%
      \end{adjustbox}
  \end{minipage}
  \hfill
  \begin{minipage}{0.37\textwidth}
    \centering
    \includegraphics[width=\columnwidth]{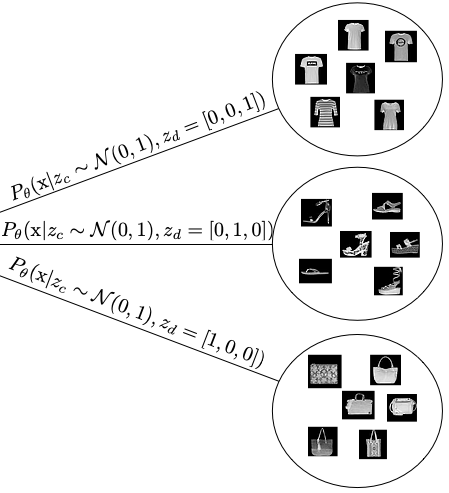}
  \end{minipage}
      \captionlistentry[table]{A table beside a figure}
    \captionsetup{labelformat=andtable}
    \caption{FashionMNIST with $L=3$ tasks: \emph{t-shirts},
      \emph{sandals} and \emph{bag}. To generate samples from the $i$-th task conditional,
      $P_i(\vect{x}) = P(\bm{\mathrm{x}} | \bm{z}_d=i)$, we set
      $\bm{z}_d = i$, randomly sample $\bm{z}_c \sim \mathcal{N}(0,
      1)$, and run $[\bm{z}_c, \bm{z}_d]$ through the decoder, $P_{\bm{\theta}}(\vect{x}|\bm{z}_c, \bm{z}_d)$. Resampling $\bm{z}_c$, while keeping $\bm{z}_d$ fixed,
      enables generation of varied samples from the task
      conditional. \emph{Left}: Desired task conditionals. \emph{Right}: Desired decoder behavior.} \label{table_gen_example}
  \end{figure}
\vskip -0.1in

Given the importance of \emph{generative replay} for knowledge
transfer in LGM, synthetic sample generation by the teacher model  needs to be representative
of all the previously observed distributions in order to prevent
catastrophic forgetting. Under the assumption that $\bm{z}_d$
accurately captures the underlying discriminativeness of the
individual distributions and through the definition of the LGM generative
process, shown in Equation \ref{gen_process}:
\vskip -0.2in
\begin{align}
    P_{\bm{\Theta}}(\bm{\mathrm{x}} | \bm{z}_d, \bm{z}_c),\
  \bm{z}_c\sim \mathcal{N}(0, 1),\ \bm{z}_d \sim Cat(1/L), \label{gen_process}
\end{align}



\noindent
\noindent we can control generations by setting a fixed value, $\bm{z}_d = i$, and randomly sampling the continuous prior, $\bm{z}_c \sim
\mathcal{N}(0, 1)$. This is possible because the trained decoder approximates the task
conditional from Section \ref{sec_lifelong_generative_modeling}:
\vskip -0.2in
\begin{align}
  P(\vect{x}|\bm{z}_d=i) = P_i(\vect{x}) &\approx
                                           \frac{P_{\bm{\Theta}}(\vect{x}|\bm{z}_c,
                                           \bm{z}_d=i)
                                           P(\bm{z}_c)}{Q_{\bm{\Phi}}(\bm{z}_c
                                           | \vect{x})}
\end{align}
\noindent where sampling the true task conditional,
$P_i(\vect{x})$, can be approximated by sampling $\bm{z}_c \sim
P(\bm{z}_c) = \mathcal{N}(0, 1)$, keeping $\bm{z}_d$ fixed, and
decoding the variates as shown in Equation \ref{ex_gen_eq} below:
\vskip -0.2in
\begin{align}
  \hat{\vect{x}} \sim P_{\bm{\Theta}}(\vect{x} |\bm{z}_c \sim
  \mathcal{N}(0, 1), \bm{z}_d=i). \label{ex_gen_eq}
\end{align}
 We provide a simple example of our desired behavior
for three generative tasks, $L = 3$, using Fashion MNIST in Figure \ref{table_gen_example} and Table
2 above. The assumption made up till now is that $\bm{z}_d$ accurately captures the discriminative aspects of each
distribution. However, there is no theoretical reason for the model to
impose this constraint on the latent variables. In practice, we often
observe that the decoder $P_{\bm{\theta}}(\vect{x}| \bm{z}_c,
\bm{z}_d)$ ignores $\bm{z}_d$ due to the much richer representation of
the continuous variable, $\bm{z}_c$. In the following section we
introduce a mutual information constraint that encourages the model to
fully utilize $\bm{z}_d$.

\note[Alexandros-16]{If time allows, and only if, give a figure with the new graphical model.}
\subsubsection{Information restricting regularizer}\label{information_regularizer}
%
As eluded to in the previous section, the synthetic samples observed
by the student model need to be representative of all previous
distributions.
In order to control sampling via the process described in Section
\ref{sampling}, we need to enforce that the discrete variable, $\bm{z}_d$, carries
the discriminative information about each distribution. Given our
graphical model from Figure \ref{online-model}-\emph{Left}, we observe
that there are two ways to accomplish this: maximize the information content between the discrete random
variable, $\bm{z}_d$ and the decoded $\hat{\vect{x}}$, or minimize the
information content between the continuous variable, $\bm{z}_c$ and
the decoded $\hat{\vect{x}}$. Since our graphical model and underlying
network does not contain skip connections, information from the input,
$\vect{x}$, has to flow through the latent variables $\bm{z} = [\bm{z}_c, \bm{z}_d]$ to reach the decoder. While
both formulations can theoretically achieve the same objective, we
observed that in practice, minimizing $I(\hat{\vect{x}}, \bm{z}_c)$
provided better results. We believe the reason for this is that
minimizing $I(\hat{\vect{x}}, \bm{z}_c)$ provides the model with more
subtle gradient information in contrast to maximizing $I(\hat{\vect{x}},
\bm{z}_d)$ which receives no gradient information when the value of
the $k$-th element of the categorical sample is 1.
We now formalize our mutual information regularizer, which
we derive from first principles in Equation \ref{mutinfo}:

\begin{adjustbox}{width=\columnwidth}
  \parbox{1.0\linewidth}{%
    \begin{align}
      \label{mutinfo}
  \mathbb{E}_{\vect{x} \sim P_i(\vect{x})}[I(\hat{\vect{x}}, \bm{z}_c)]
  &= \mathbb{E}_{\vect{x} \sim P_i(\vect{x})}[H(\bm{z}_c) - H(\bm{z}_c |
    \hat{\vect{x}})] \\
  &= \underbrace{\mathbb{E}_{\vect{x} \sim P_i(\vect{x})}
    \mathbb{E}_{\bm{z}_c \sim Q_{\bm{\phi}}(
    \bm{z}_c | \vect{x})} \bigg[-\log
    Q_{\bm{\phi}}(\bm{z}_c | \vect{x})\bigg]}_{\mathbb{E}_{\vect{x} \sim P_i(\vect{x})}[H(\bm{z}_c)]} \nonumber\\[-0.6em]
      &\hspace{1.3in}+ \nonumber \\[-0.8em]
      &\hspace{0.2in}\underbrace{\mathbb{E}_{\vect{x} \sim P_i(\vect{x})}
    \mathbb{E}_{(\bm{z}_d, \bm{z}_c) \sim Q_{\bm{\phi}}(\bm{z}_c,
    \bm{z}_d | \vect{x})} \bigg[\mathbb{E}_{\hat{\vect{x}} \sim
    P_{\bm{\theta}}(\hat{\vect{x}}| \bm{z}_c, \bm{z}_d)} \log
        Q_{\bm{\phi}}(\bm{z}_c | \hat{\vect{x}}) \bigg]}_{\mathbb{E}_{\vect{x} \sim P_i(\vect{x})}[-H(\bm{z}_c |
    \hat{\vect{x}})]}, \nonumber
\end{align}
}
\end{adjustbox}

\noindent where we use the independence assumption of our posterior
from Equation \ref{decompose_eqn} and the fact that the expectation of
a constant is the constant. This regularizer has parallels to the
regularizer in InfoGAN \cite{NIPS2016_6399}. In contrast to InfoGAN, VAEs already estimate the posterior $Q_{\bm{\phi}}(\bm{z}_c | \bm{\mathrm{x}})$ and thus do not need the
introduction of any extra parameters $\bm{\phi}$ for the
approximation. In addition \cite{huszar_2016} demonstrated that InfoGAN uses the variational bound (twice) on the mutual
information, making its interpretation unclear from a theoretical
point of view.
In contrast, our regularizer 
has a clear
interpretation: it restricts information through a specific latent variable within the computational graph.
We observe that this constraint is essential for empirical performance
of our model and empirically validate this in our ablation study in
Experiment \ref{ablation}.

\subsection{Learning Objective}\label{full_learning_objective}

The final learning objective for each of the student models is the maximization
of the sequential VAE ELBO (Equation \ref{student_elbo_bayes_eqn}), coupled with generative replay (Equation \ref{xsampling})
and the mutual information regularizer, $I(\hat{\vect{x}}, \bm{z}_c)$,
(Equation \ref{mutinfo}):



\begin{center}
\begin{flalign}
  \label{eqn7}
  \mathcal{L}_{\bm{\theta}, \bm{\phi}}(\vect x) &= \
  \underbrace{\mathbb{E}_{Q_{\bm{\phi}}(\bm{z}_c, \bm{z}_d|\vect x)} \bigg[ \log
     P_{\bm{\theta}}(\vect{x}| \bm{z}_c, \bm{z}_d) \bigg] -
                                                         \text{KL}
     [Q_{\bm{\phi}}(\bm{z}_c, \bm{z}_d|\vect{x}) || P(\bm{z}_c, \bm{z}_d)]}_{\text{VAE ELBO}} \\ \nonumber
   & \hspace{18.5mm}+ \underbrace{(1-\omega)\text{KL}[
     Q_{\bm{\phi}}(\bm{z}_c, \bm{z}_d |    {\bm{\mathrm{x}}}     ) ||
     Q_{\bm{\Phi}}(\bm{z}_c, \bm{z}_d
     |    {\bm{\mathrm{x}}}    ) ]}_{\text{Posterior Consistency Regularizer}}  \nonumber \\
  & \hspace{35.5mm} - \underbrace{\lambda I(\hat{\bm{\mathrm{x}}},
    \bm{z}_c)}_{\text{Mutual Information}}, \nonumber \\
& \hspace{30mm} \bm{\mathrm{x}} \sim P(\bm{\omega})P(\bm{\mathrm{x}} |
\bm{\omega}), \, \bm{\omega} \sim \text{Ber}(\pi) \nonumber
\end{flalign}
\end{center}

The $\lambda$
hyper-parameter controls the importance of the information gain
regularizer.  Too large a value for $\lambda$ causes a lack of  sample
diversity, while too small a value causes the model to not use the
discrete latent distribution. We did a random hyperparameter search
and determined $\lambda = 0.01$ to be a reasonable choice for all of
our experiments. This is in line with the $\lambda$ used in InfoGAN
\cite{NIPS2016_6399} for continuous latent variables. We empirically validate the
necessity of both terms proposed in Equation \ref{eqn7} in our
ablation study in Experiment \ref{ablation}.  We also validate the benefit of the
latent variable factorization in Experiment \ref{experiment0}. Before
delving into the experiments, we provide a theoretical
analysis of computational complexity induced by our model and objective (Equation \ref{eqn7}) in Section \ref{comp_complex_sec} below.





\subsection{Computational Complexity}\label{comp_complex_sec}

We define the computational complexity of a typical VAE encoder and
decoder as $O(E)$ and $O(D)$ correspondingly; internally these are
dominated by the matrix-vector products which take approximately $L O(n^2)$ for L layers. We also define
the cost of applying the loss function as $O(K) + O(R)$, where $O(K)$
is the cost of evaluating the KL divergence from the ELBO (Equation
\ref{elbo_eqn}) and $O(R)$ the cost for evaluating the
reconstruction term. Given these definitions, we can summarize LGM's computation complexity as follows in Equation \ref{comp_complex_eqn}:
\begin{adjustbox}{width=1.0\linewidth}
  \parbox{1.0\linewidth}{%
\begin{align}
  &\underbrace{O(D)}_{\substack{\text{teacher} \\ \text{generations}}} + \underbrace{O(E) +
  O(D)}_{\substack{\text{student encode} \\ \text{+ decode}}} + \underbrace{O(K) +
  O(R)}_{\text{vae loss}} +
  \underbrace{O(K)}_{\substack{\text{posterior} \\
  \text{regularizer}}} + \underbrace{O(K) + O(E)}_{\text{mutual info}} \nonumber \\
  &= 2 [O(D) + O(E)] + 3[O(K)] + O(R), \label{comp_complex_eqn}
\end{align}
}
\end{adjustbox}

\noindent where we introduce increased computational complexity due to
teacher generations, the cost of the posterior regularizer, and the mutual information
terms; the latter of which necessitates an extra encode operation,
$O(E)$. The computational complexity is still dominated by the matrix-vector product
from evaluating forward functionals of the neural network. These operations can easily be amortized through parallelization on modern
GPUs and typical experiments do not directly scale as per Equation \ref{comp_complex_eqn}. In our most demanding experiment
(Experiment \ref{experiment4}), we observe an average empirical increase of 13.53
seconds per training epoch and 6.3 seconds per test epoch.


%% file: related_plus_plus.tex
\section{Revisiting state of the art methods.}

In this section we revisit some of the state of the art methods from
Section \ref{rel_sec}. We begin by providing a mathematical
description of the differences between EWC
\cite{kirkpatrick2017overcoming}, VCL \cite{nguyen2018variational} and LGM and follow it up
with a discussion of VASE \cite{achille2018life} and their extensions of our work.

\textbf{EWC and VCL}: \label{EWC_VS_LL}
Our posterior regularizer, $KL[ Q_{\bm{\phi}}(\bm{z} |    {\bm{\mathrm{x}}}     ) || Q_{\bm{\Phi}}(\bm{z}
  |    {\bm{\mathrm{x}}}    ) ]$, affects the same parameters, $\bm{\phi}$, as parameter regularizer methods such as
EWC and VCL. However, rather than assuming a functional form for the
parameter posterior, $P(\bm{\phi} | \vect{x})$, our method regularizes
the \emph{output} latent distribution $Q_{\bm{\phi}}(\bm{z} |
\vect{x})$. EWC and VCL, both make the assumption that $P(\bm{\phi} | \vect{x})$ is distributed
as an isotropic gaussian\footnote{VCL assumes an isotropic gaussian
  variational form vs. EWC which directly assumes the parametric form
on $P(\bm{\phi}|\vect{x})$.}. This allows the use of the Fisher
Information Matrix (FIM) in a quadratic parameter regularizer in EWC, and an analytical KL divergence of the
posterior in VCL. This is a very stringent requirement for the
parameters of a neural network and there is active research in Bayesian neural
networks that attempts to relax this constraint \cite{louizos2016structured,louizos2017multiplicative,mishkin2018slang}.

\begin{center}
\scalebox{0.7}{
\begin{tabular}{ c | c } \hline
  EWC $\ \ \ \min_{\bm{\phi}} d[P(\bm{\phi} | \bm{\mathrm{x}}) || P(\bm{\Phi} |
  \bm{\mathrm{x}})]$  & LGM ( Isotropic Gaussian Posterior ) $\ \ \
               \min_{\bm{\phi}} d[Q_{\bm{\phi}}(\bm{z} | \bm{\mathrm{x}}) || Q_{\bm{\Phi}}(\bm{z} | \bm{\mathrm{x}})]$\\ \hline
$  \approx \frac{\gamma}{2}
        (\bm{\phi} - \bm{\Phi})^T F (\bm{\phi} - \bm{\Phi})$ &
                                                               $ =
                                                               0.5\bigg[tr(\bm{\Sigma_{\bm{\Phi}}}^{-1} \bm{\Sigma_{\bm{\phi}}}) + (\bm{\mu_{\bm{\Phi}}} - \bm{\mu_{\bm{\phi}}})^T \bm{\Sigma_{\bm{\Phi}}}^{-1}(\bm{\mu_{\bm{\Phi}}} - \bm{\mu_{\bm{\phi}}}) - C + log \bigg( \frac{|\bm{\Sigma_{\bm{\Phi}}}|}{|\bm{\Sigma_{\bm{\phi}}}|} \bigg)\bigg]$
\end{tabular}\label{contrast}}
\end{center}

In the above table 
we examine the distance metric $d$, used to minimize the
effects of catastrophic inference in both EWC and LGM. While our method can operate over any distribution
that has a tractable KL-divergence, for the purposes of
demonstration we examine the simple case of an isotropic gaussian
latent-variable posterior. EWC directly enforces a quadratic constraint on
the model parameters $\bm{\phi}$, 
while our method
indirectly affects the same parameters through a regularization of the
posterior distribution $Q_{\bm{\phi}}(\bm{z}|\bm{\mathrm{x}})$. For
any given input variate, $\bm{\mathrm{x}}_j$, LGM allows to model to freely change its
internal parameters, $\bm{\phi}$; it does so in a
non-linear\footnote{This is because the parameters of the distribution
are modeled by a deep neural network.} way such that the analytical KL shown above is
minimized.

\textbf{VASE} : The recent work of Life-Long Disentangled Representation Learning with
Cross-Domain Latent Homologies (VASE) \cite{achille2018life} extend upon our
work \cite[p.~7]{achille2018life}, but take a more empirical route by
incorporating a classification-based heuristic for their posterior
distribution. In contrast, we show (Section \ref{regularizer}) that
our objective naturally emerges in a sequential learning setting for
VAEs, allowing us to infer the discrete posterior,
$Q_{\bm{\phi}}(\bm{z}_d | \vect{x})$ in an unsupervised
manner. Due to the incorporation of direct supervised class information
\cite{achille2018life} also observe that regularizing the decoding
distribution $P_{\bm{\theta}}(\bm{\mathrm{x}} | \bm{z})$ aids in the learning
process, something that we observe to fail in a purely
unsupervised generative setting (Appendix Section
\ref{ll_regularizer}). Finally, in contrast to \cite{achille2018life},
we include an information restricting regularizer (Section \ref{information_regularizer}) which
allows us to directly control the interpretation and flow of
information of the learnt latent variables.


%% file: experiments.tex
\section{Experiments}\label{experiments}



We evaluate our model and the baselines over \emph{standard datasets} used in
other state of the art lifelong / continual learning literature
\cite{nguyen2018variational,
  zenke2017continual,shin2017continual,kamra2017deep,kirkpatrick2017overcoming,rusu2016progressive}. While
these datasets are simple in a traditional classification setting, transitioning
to a \emph{lifelong-generative} setting scales the problem complexity
substantially. We evaluate LGM on a set of progressively more
complicated tasks (Section \ref{data_flow_sec}) and provide
comparisons against baselines
\cite{nguyen2018variational,zenke2017continual,kirkpatrick2017overcoming,eskin2004laplace,kingma2014}
using a set of standard metrics (Section
\ref{metrics_sec}). All \emph{network architectures} and other
optimization details for our LGM model are provided in Appendix
Section \ref{arch} as well our open-source git repository
\cite{jramapuram_2018}.

\subsection{Performance Metrics} \label{metrics_sec}
To validate the benefit of LGM in a lifelong setting we explore three
main performance dimensions: the ability for the model to
\emph{reconstruct} and \emph{generate} samples from all previous tasks
and the ability to learn a common representation over time, thus
\emph{reducing learning sample complexity}.
We use three main quantitative performance metrics for our experiments:
 the log-likelihood importance sample estimate \cite{burda2015importance,nguyen2018variational}, the \emph{negative} test ELBO, and the Frechet distance metric \cite{heusel2017gans}. In addition, we also provide two auxiliary
metrics 
to validate the benefits of LGM
in a lifelong setting: training sample complexity and wall clock time per training and test epoch.

To fairly compare models with varying latent variable configurations, one solution is to marginalize out the
latents, $\bm{z}$, during model evaluation / test time: $\int_{\bm{z}}
P_{\bm{\theta}}(\vect{x} | \bm{z}) d\bm{z} \approx \sum_{k=1}^K
P_{\bm{\theta}}(\vect{x} | \bm{z} = \bm{z}_{k})$. This is realized in practice by using a
Monte Carlo approximation (typically K=5000) and is commonly known as the importance
sample (IS) log-likelihood estimate \cite{burda2015importance,nguyen2018variational}. As latent variable
and model complexity grows, this estimate tends to become noisier and
intractable to compute. For our experiments we use this metric
only for the FashionMNIST and MNIST datasets as computing one estimate
over 10,000 test samples for a complex model takes approximately 35 hours on a K80 GPU.

In contrast to the IS log-likelihood estimate, the negative test ELBO (Equation \ref{elbo_eqn}) is only
applicable when comparing models with the same latent variable
configurations; it is however much faster to compute. The negative test ELBO provides a lower bound
to the test log-likelihood of the true data distribution under the
assumed latent variable configuration. One crucial aspect missing from
both these metrics is an evaluation of generation quality. We
resolve this by using the Frechet distance metric
\cite{heusel2017gans} and qualitative image samples.

The Frechet distance metric allows us to quantify the quality and
diversity of generated samples by using a pre-trained classifier model to compare the feature statistics (generally under a
Gaussianity assumption) between synthetic generated samples and samples drawn from the test set. If the Frechet distance
between these two distributions is small, then the generative model
is said to be generating realistic images. The Frechet
distance between two gaussians (produced by evaluating latent embeddings of a classifier model) with means $\bm{m}_{test}, \bm{m}_{gen}$ with
corresponding covariances $\bm{C}_{test}, \bm{C}_{gen}$ is:
\vskip -0.2in
\begin{align}
||\bm{m}_{test} - \bm{m}_{gen}||_2^2 + Tr(\bm{C}_{test} + \bm{C}_{gen} - 2
  [\bm{C}_{test}\bm{C}_{gen}]^{0.5}).\label{frechet_eqn}
\end{align}

While the Frechet distance, negative ELBO and IS log-likelihood
estimate provide a glimpse into model performance, there
exists no conclusive metric that captures the quality of unsupervised
generative models \cite{theis2016note, sajjadi2018assessing} and active research suggests a
direct trade-off between perceptual quality and model representation \cite{blau2018perception}.
Thus, in addition to the metrics described above, we also provide
qualitative metrics in the form of test image reconstructions and image generations. We summarize all used performance metrics in Table \ref{metrics_table} below:

\vskip -0.1in
\begin{table}[H]
        {\renewcommand{\arraystretch}{1.2}%
        \begin{center}
          \begin{adjustbox}{width=\columnwidth}

\begin{tabular}{l|l|l|l|}
\cline{2-4}
                                                    & \textbf{Definition}                                                                          & \textbf{Purpoose}                                                                               & \textbf{Lower is better?} \\ \hline
\multicolumn{1}{|l|}{\textbf{Negative ELBO}}        & Equation \ref{elbo_eqn}.                                                                                          & \begin{tabular}[c]{@{}l@{}}Quantitative metric on \\ likelihood / reconstructions.\end{tabular} & yes                       \\ \hline
\multicolumn{1}{|l|}{\textbf{Negative Log-Likelihood}}
                                                    & \begin{tabular}[c]{@{}l@{}}5000 (latent) sample Monte \\ Carlo estimate of  Equation \ref{elbo_eqn}.\end{tabular} & \begin{tabular}[c]{@{}l@{}}Quantitative metric on \\ density estimate.\end{tabular}             & yes                       \\ \hline
\multicolumn{1}{|l|}{\textbf{Frechet Distance}}     & Equation \ref{frechet_eqn}.                                                                                           & Quantitative metric on generations.                                                             & yes                       \\ \hline
\multicolumn{1}{|l|}{\textbf{Test Reconstructions}} & $P_{\bm{\theta}}(\vect{x}
                                                      | \bm{z}_c \sim
                                                      Q_{\bm{\phi}}(\bm{z}_c|\vect{x}),
                                                      \bm{z}_d\sim
                                                      Q_{\bm{\phi}}(\bm{z}_d
                                                      | \vect{x}))$                                                                                     & Qualitative view of reconstructions.                                                            & N/A                       \\ \hline
\multicolumn{1}{|l|}{\textbf{Generations}}          & $P_{\bm{\theta}}(\vect{x}|\bm{z}_d \sim Cat(1/L), \bm{z}_c \sim \mathcal{N}(0, 1))$.                       & Qualitative view of generations.                                                                & N/A                       \\ \hline
\multicolumn{1}{|l|}{\textbf{\#Training Samples}}   & \# real training samples used for task $i$.                                                    & Sample Complexity.                                                                              & yes                       \\ \hline
\end{tabular}
\end{adjustbox}
\end{center}
}
\caption{Summary of different performance
  metrics.}\label{metrics_table}
\vskip -0.3in
\end{table}

\subsection{Data Flow} \label{data_flow_sec}
\begin{figure}[H]
  \includegraphics[width=\columnwidth]{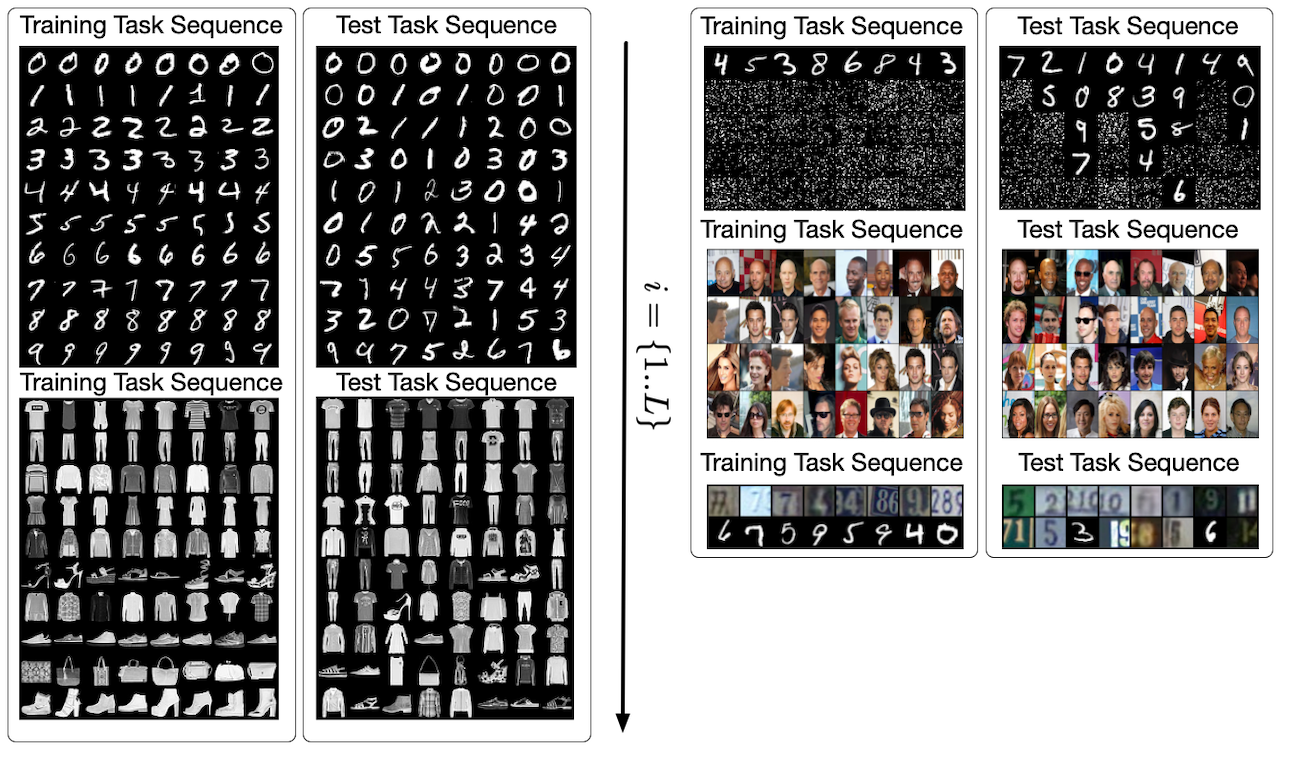}
  \caption{Visual examples of training and test task sequences (top to bottom) for
    the datasets used to validate LGM. The training set only consists of
    samples from the current task while the test set is a cumulative union of the current task, coupled with all previous
    tasks. The permuted MNIST tasks uses $\{G_1, ... G_{L-1}\}$ different fixed
    permutation matrices to create 4 auxiliary datasets.}\label{data_flow_fig}
\end{figure}

In Figure \ref{data_flow_fig} we list train and test variates depicting the data flow for each of the
problems that we model. Due to the relaxing the need to preserve data in
a lifelong setting, the train task sequence observes a \emph{single
dataset}, $\bm{\mathcal{D}}^{tr}_i$, at a time, without access to any
previous, $\bm{\mathcal{D}}^{tr}_{<i}$. The corresponding test dataset consists of a union ($\cup$ operator) of the current test
dataset, $\bm{\mathcal{D}}^{te}_{i}$, merged with all previously observed test datasets,
$\hat{\bm{\mathcal{D}}}^{te}_{i} = \bm{\mathcal{D}}^{te}_{i} \cup
\bm{\mathcal{D}}^{te}_{i-1} \cup ... \cup
\bm{\mathcal{D}}^{te}_{1}$.

\noindent \textbf{MNIST / Fashion MNIST}: For the MNIST and Fashion MNIST problems, we observe
a single MNIST digit or fashion object (such as shirts) at a time. Each training set consists of 6000 training samples and 1000 test samples. These
samples are originally extracted from the full training and test
datasets which consist of 60,000 training and 10,000 test samples.

\noindent \textbf{Permuted MNIST}: this problem differs from the MNIST
problem described above in that we use
the entire MNIST dataset at each task. After observing the first task,
which is the standard MNIST dataset, each subsequent task differs through
the application of a fixed permutation matrix $G_i$ on the entire
MNIST dataset. The test task sequence differs from the training task
sequence in that we simply use the corresponding full train and test
MNIST datasets (with the appropriate application of $G_i$).

\noindent \textbf{Celeb-A}: We split the CelebA dataset into four individual distributions
using the features:  \emph{bald}, \emph{male}, \emph{young} and \emph{eye-glasses}
. As with the previous problems, we treat each
subset of data as an individual distribution, and present our model samples
from a single distribution at a time. This presents a real world
scenario as the samples per distribution varies drastically from
only 3,713 samples for the \emph{bald} distribution, to 126,788 samples
for \emph{young}. In addition specific samples can span one or more of these
distributions.

\noindent \textbf{SVHN to MNIST}:
in this problem, we transition from
fully observing the centered SVHN \cite{netzer2011reading} dataset to observing the MNIST
dataset. We treat all samples from
SVHN as being generated by one distribution $P_1(\bm{\mathrm{x}})$ and all the
MNIST \footnote{MNIST was resized to 32x32 and converted to RBG to make it consistent with the dimensions of SVHN.}
samples as generated by another distribution $P_2(\bm{\mathrm{x}})$
(irrespective of the specific digit). At inference, the model is
required to reconstruct and generate from both datasets.

\vspace{-0.1in}
\subsection{Situating against state of the art lifelong learning
  models.}
\vspace{-0.1in}

To situate LGM against other state of the art methods in
lifelong learning we use the sequential FashionMNIST and MNIST
datasets described earlier in Section \ref{data_flow_sec} and the data
flow diagram in Figure \ref{data_flow_fig}.
We contrast our LGM model against VCL \cite{nguyen2018variational},
VCL without a task specific head network, SI
\cite{zenke2017continual}, EWC \cite{kirkpatrick2017overcoming},
Laplace propagation \cite{eskin2004laplace}, a full \emph{batch} VAE
trained jointly on all data and a standard \emph{naive}
sequential VAE without any catastrophic forgetting prevention strategy in Figures
\ref{importance_fig} and \ref{mnist_all} below. The full \emph{batch} VAE presents
the upper-bound performance and all lifelong learning models typically
under-perform this model by the final learning task. For the baselines, we use the generously
open sourced code \cite{nvcuong_2019} by the VCL authors, using the optimal hyper-parameters
specified for each model. We begin by evaluating the 5000 sample Monte Carlo estimate of the log-likelihood of all
compared models in Figure \ref{importance_fig} below:

\begin{figure}[H]
\begin{center}
  \scalebox{1.0}{
\minipage{0.49\textwidth}
\includegraphics[width=\linewidth]{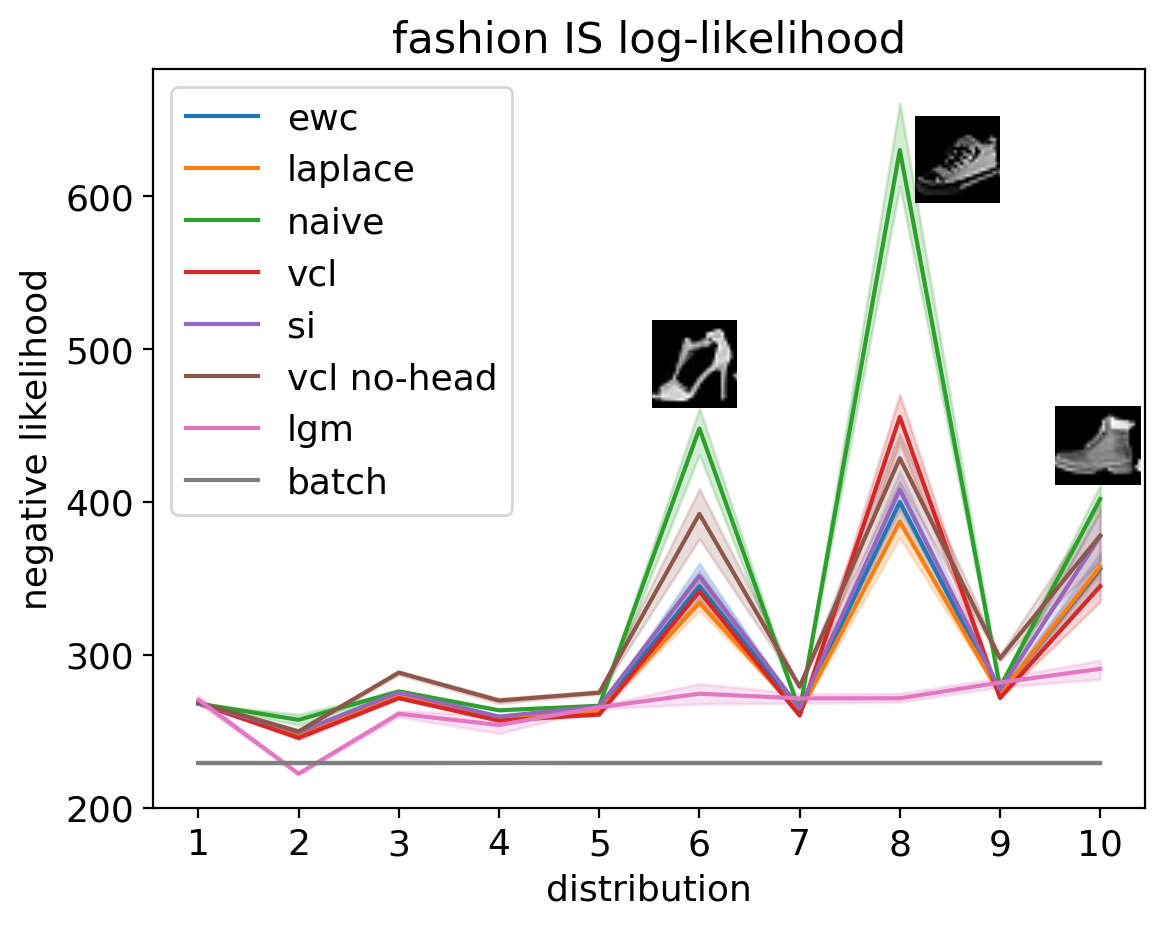}
    \endminipage\hfill
  }\scalebox{1.0}{
\minipage{0.49\textwidth}
\includegraphics[width=\linewidth]{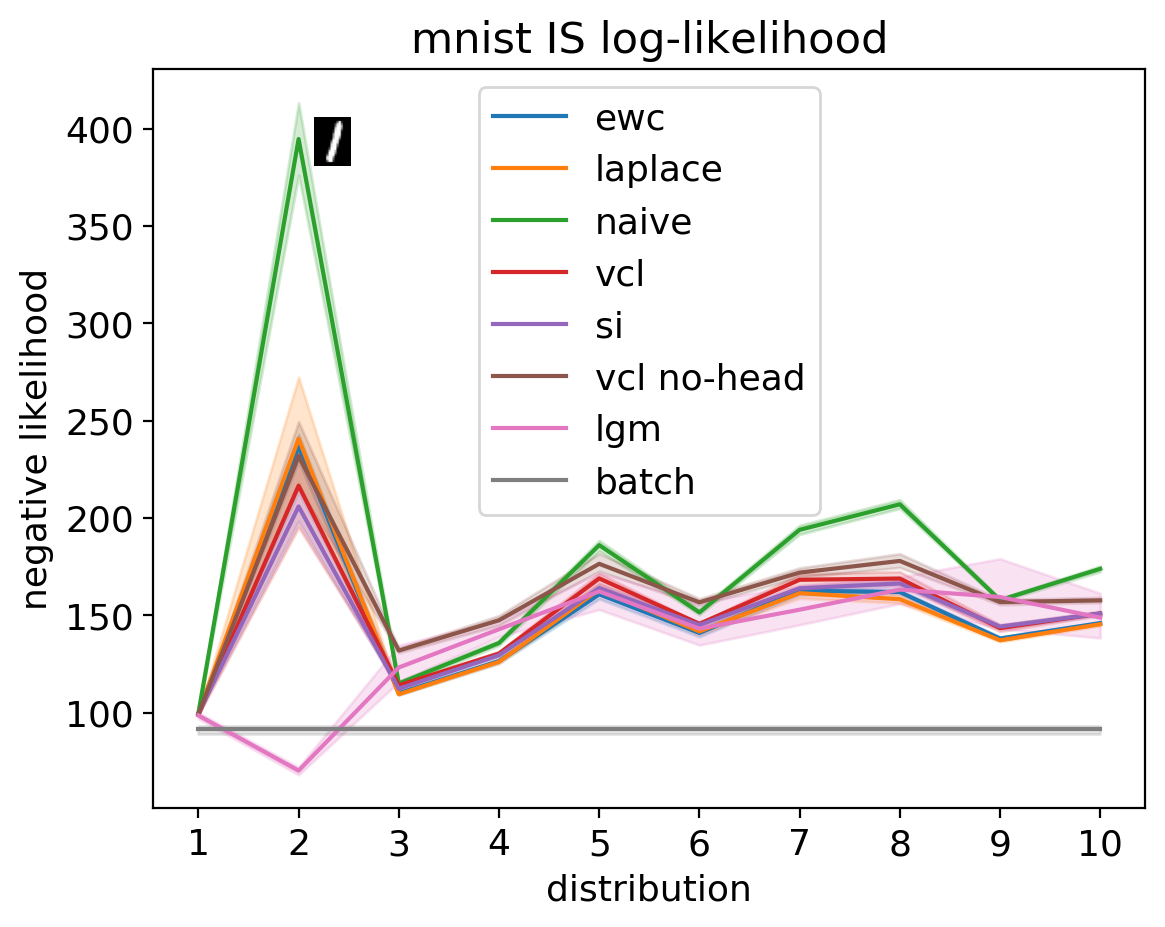}
\endminipage%
}
\end{center}%
\vskip -0.2in
\caption{IS log-likelihood (mean $\pm$ std) $\times$ 5. \emph{Left}: Fashion
  MNIST. \emph{Right}: MNIST.} \label{importance_fig}
\end{figure}%
\vspace{-0.3in}
Even though each trial was repeated five times (each), we observe
large increases in the estimates at a few critical points. After
further inspection, we determined the large magnitude increases were
due to the model observing a drastically different distribution at
that point. We overlay the graphs with an example variate for of the
magnitude spikes. In the case of FashionMNIST for example, the model
observes its first shoe distribution at $i=6$; this contrasts the
previously observed items which were mainly clothing related
objects. Interestingly we observe that LGM has much smoother
performance across tasks. We posit this is because LGM does not
constrain its parameters, and instead enforces the same input-output
mapping through \emph{functional regularization}.

\vskip -0.1in
\begin{figure}[H]
\begin{adjustbox}{width=\columnwidth}
  \includegraphics[width=\columnwidth]{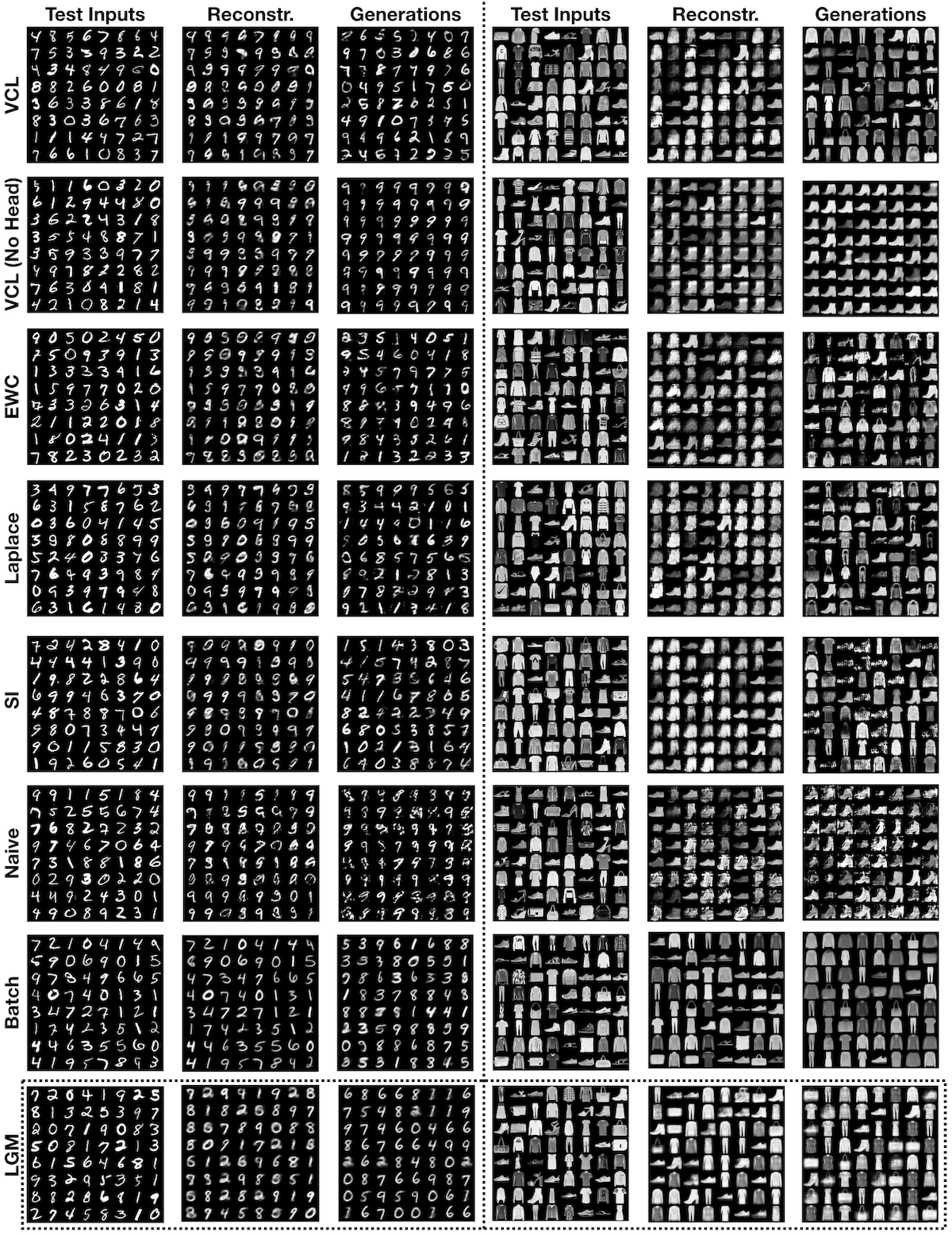}
\end{adjustbox}
\vskip -0.1in
\caption{Final model, $i=10$, generation and reconstructions for
    MNIST and FashionMNIST. The LGM model presents competitive
    performance for both generations and reconstructions, while not
    preserving any past data nor past models.}\label{mnist_all}
\end{figure}

\begin{figure}[H]
\begin{center}
  \scalebox{1.0}{
\minipage{0.49\textwidth}
\includegraphics[width=\linewidth]{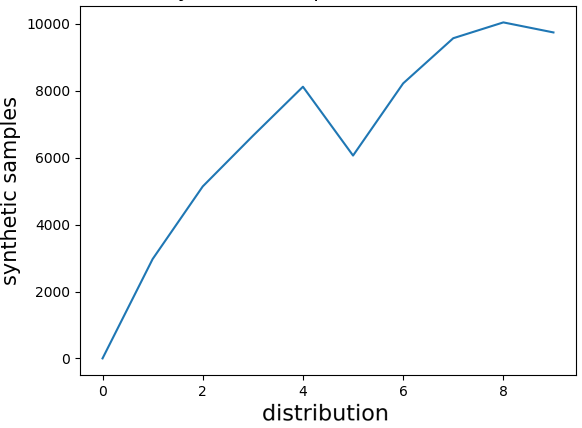}
    \endminipage\hfill
  }\scalebox{1.0}{
\minipage{0.49\textwidth}
\includegraphics[width=\linewidth]{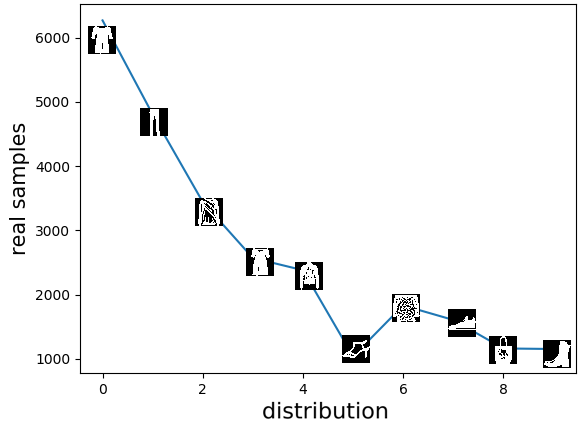}
\endminipage%
}
\end{center}%
\vskip -0.2in
\caption{FashionMNIST sample complexity. \emph{Left}: Synthetic training samples used
   till early-stopping. \emph{Right}: Real samples used till early-stopping.} \label{sample_complexity_fig}
\vskip -0.1in
\end{figure}

Since one of the core tenets of lifelong learning is to reduce sample
complexity over time, we use this experiment to validate if LGM does
in fact achieve this objective.
Since all LGM models are trained with an early-stopping criterion, we
can directly calculate the number of samples used for each learning
task using the stopping epoch and mean, $\pi$ of the Bernoulli
sampling distribution of the student model. In Figure \ref{sample_complexity_fig} we plot the number
of true samples and the number of synthetic samples used by a model until
it satisfied its early-stopping criterion. We observe a steady decrease
in the number of real samples used over time, validating LGMs
advantage in a lifelong setting.

\vskip -0.15in

\vspace{-0.2in}
\subsection{Diving deeper into the sequence.}
\begin{figure}[H]
\vskip 0.2in
  \begin{center}
    \centerline{\includegraphics[width=\columnwidth]{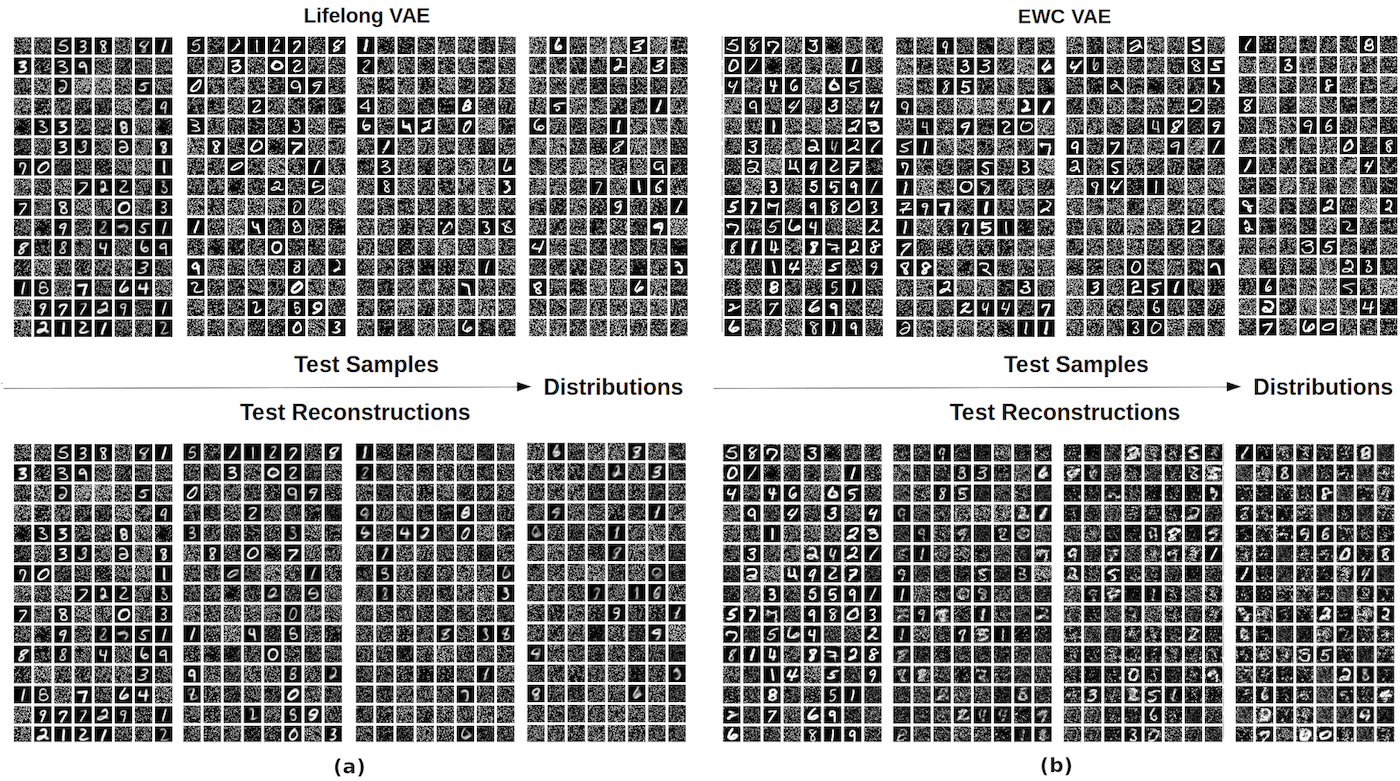}}
    \vskip -0.1in
\caption{Top row: test-samples; bottom row: reconstructions. We
  visualize an increasing number of accumulated distributions from left to right. (a)
  Lifelong VAE model (b) EWC VAE
  model.}\label{fig_fashion_permute_reconstr}
\end{center}
\end{figure}

\vspace{-0.1in}
Rather than only visualizing the final model's qualitative results as in
Figure \ref{mnist_all}, we provide qualitative results for model
performance over time for the PermutedMNIST experiment in Figure \ref{fig_fashion_permute_reconstr}. This allows us
to visually observe lifelong model performance over time.
In this experiment, we focus our efforts on EWC and LGM and visualize model
(test) reconstructions starting from the second learning task, $G_1
\bm{\mathcal{D}}$, till the final $G_4 \bm{\mathcal{D}}$. The EWC-VAE
variant that we use as a baseline has the same latent variable
configuration as our model, enabling the usage of the test ELBO as a
quantitative metric for comparison. We use an unpermuted version of the MNIST dataset, $\bm{\mathcal{D}}$,
as our first distribution, $P_1(\bm{\mathrm{x}})$, as it allows us to
visually asses the degradation of reconstructions.
This is a common setup utilized in continual learning \cite{kirkpatrick2017overcoming, zenke2017continual} and we extend it here
to the density estimation setting. 

\vskip -0.2in
\vskip -0.2in

Both models exhibit a different form of degradation: EWC experiences a
more destructive form of degradation as exemplified by the
salt-and-pepper noise observed in the final dataset reconstruction at $G_4
\bm{\mathcal{D}}$. LGM on the hand experiences a form of Gaussian
noise as visible in the corresponding final dataset reconstruction. In
order to numerically quantify this performance we analyze the
\emph{log}-Frechet distance and negative ELBO below in Figure \ref{permuted},
where we contrast the LGM to EWC, a batch VAE (\emph{full-vae} in graph),
an \emph{upto}-VAE that observes all training data up to the current
distribution and a vanilla sequential VAE (\emph{vanilla}). We examine a variety of
different convolutional and dense architectures and present the top
performing models below. We observe that LGM drastically outperforms EWC and the baseline naive
sequential VAE in both metrics.

\begin{figure}[H]
\begin{center}
  \includegraphics[width=\linewidth]{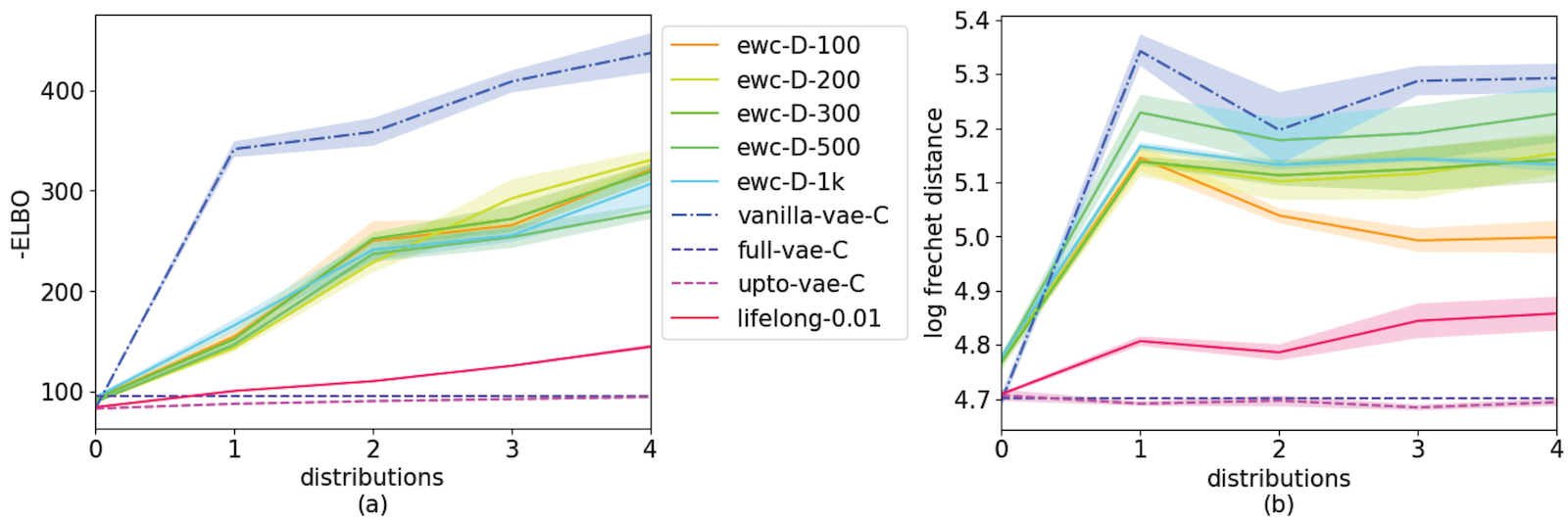}
\end{center}%
\caption{PermutedMNIST (a) negative Test ELBO and (b)
  \emph{log}-Frechet distance.} \label{permuted}
\end{figure}%

\subsection{Learning Across Complex Distributions} \label{experiment3}



\begin{figure}[H]
\begin{center}
  \scalebox{1.0}{\centerline{\includegraphics[height=40mm]{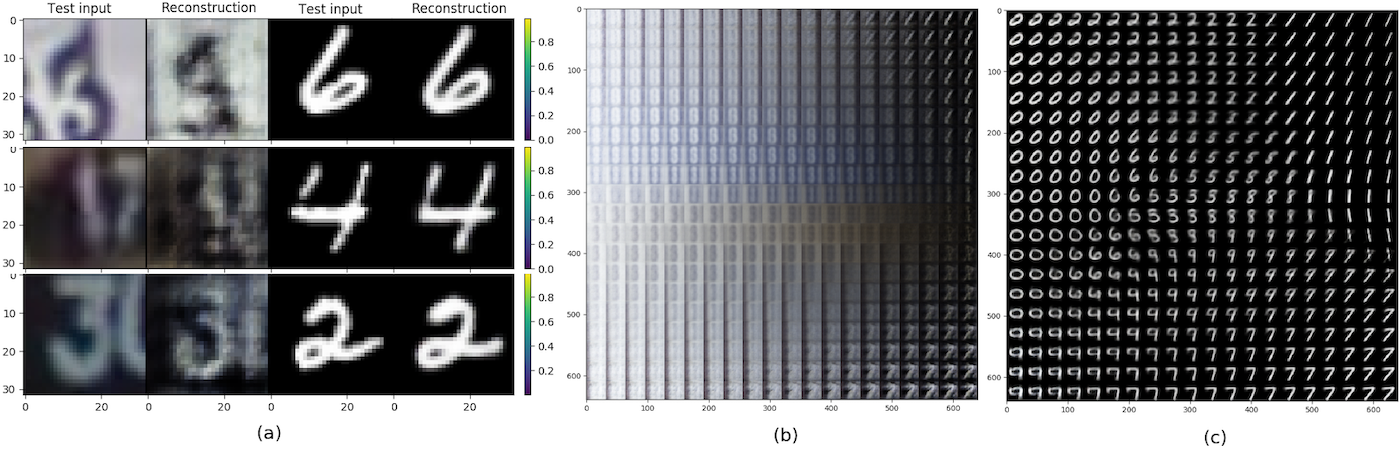}}}
\caption{(a) Reconstructions of test samples from SVHN[left] and MNIST[right]; (b) Decoded samples $\hat{\bm{\mathrm{x}}} \sim P_{\bm{\theta}}(\bm{\mathrm{x}} | \bm{z}_d, \bm{z}_c)$ based on linear interpolation of $\bm{z}_c \in \mathcal{R}^2$ with $\bm{z}_d = [0,1]$; (c) Same as (b) but with $\bm{z}_d = [1,0]$.}
\label{fig_svhn}
\end{center}
\end{figure}


The typical assumption in lifelong learning is that the sequence of
observed distributions are related \cite{thrun1995lifelong2} in some
manner. In this experiment we relax this constraint by learning a common model
between the colored SVHN dataset and the binary MNIST dataset. While
semantically similar to humans, these datasets are vastly different, as
one is based on RGB images of real world house numbers and the other of
synthetically hand-drawn digits. We visualize examples of the true
test inputs, $\bm{\mathrm{x}}$, and their respective reconstructions, $\hat{\bm{\mathrm{x}}}$, from the final
lifelong model in figure \ref{fig_svhn}(a).
Even though the only true data the final model received for training
was the MNIST dataset, it is still able to reconstruct the SVHN data
observed previously. This demonstrates the ability of our architecture to transition between complex
distributions while still preserving the knowledge learned from the previously
observed distributions.

Finally, in figure \ref{fig_svhn}(b) and \ref{fig_svhn}(c) we illustrate the data generated from an interpolation of a 2-dimensional
continuous latent space, $\bm{z}_c \in \mathcal{R}^2$. To generate
variates, we set the discrete categorical, $\bm{z}_d$, to one of the possible values
$\{[0,1],[1,0]\}$ and linearly interpolate the continuous $\bm{z}_c$ over the
range $[-3, 3]$. We then decode these to
obtain the samples, $\hat{\bm{\mathrm{x}}} \sim P_{\bm{\theta}}(\bm{\mathrm{x}} |
\bm{z}_d, \bm{z}_c)$. The model learns a common continuous structure for the two distributions which
can be followed by observing the development in the generated samples from top
left to bottom right on both figure \ref{fig_svhn}(b) and \ref{fig_svhn}(c).

\subsection{Validating Empirical Sample Complexity Using Celeb-A} \label{experiment4}

\begin{figure*}[ht]
\begin{minipage}{0.48\textwidth}
\includegraphics[width=\linewidth]{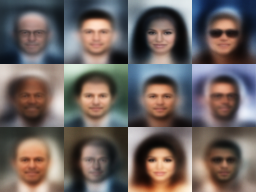}
\end{minipage}%
\hfill%
\begin{minipage}{0.47\textwidth}
\includegraphics[width=\linewidth]{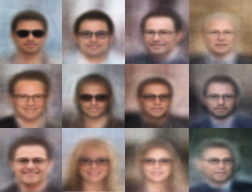}
\end{minipage}
\caption{\emph{Left}: Sequential generations for Celeb-A from the
  final \emph{lifelong} model for \emph{bald}, \emph{male},
  \emph{young} and \emph{eye-glasses} (left to
  right). \emph{Right}: (random) generations by the final baseline VAE
  model.}\label{celeba_exp}
\end{figure*}

\begin{figure}[H]
\begin{minipage}{0.4\textwidth}
  \includegraphics[width=\linewidth]{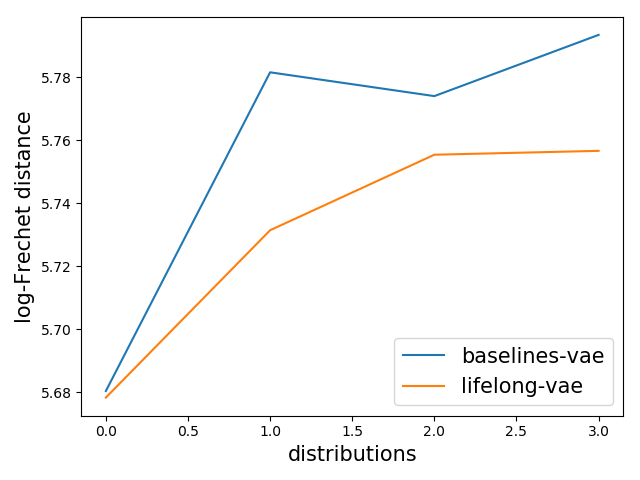}
\end{minipage}%
\hfill%
\begin{minipage}{0.6\textwidth}
\begin{table}[H]
  \begin{center}
    \begin{adjustbox}{width=\columnwidth}
\begin{tabular}{l|l|l|}
\cline{2-3}
44,218 \emph{male} samples                     & baseline-VAE  & Lifelong        \\ \hline
\multicolumn{1}{|l|}{training-epoch (s)} & 43.1 +/- 0.6 & 56.63 +/- 0.28  \\ \hline
\multicolumn{1}{|l|}{testing-epoch  (s)} & 9.79 +/- 0.12 & 16.09 +/- 0.01 \\ \hline
\end{tabular}
\end{adjustbox}
\caption{Mean \& standard deviation wall-clock for one epoch of \emph{male}
  distribution of Celeb-A.}\label{wall_clock}
\end{center}
\end{table}
\end{minipage}%
\caption{Celeb-A \emph{log}-Frechet distance of \emph{lifelong}
  vs. naive \emph{baseline} VAE model without catastrophic mitigation
  strategies over the four distributions. Listed on the
  right is the time per epoch (in seconds) for an epoch of the corresponding models. }\label{ca_fid}
\end{figure}

We iterate the Celeb-A dataset as described in the data flow diagram (Figure \ref{data_flow_fig}) and use
this learning task to explore qualitative and quantitative
generations, as well as empirical real world time complexity (as described in Section
\ref{comp_complex_sec}) on modern GPU hardware. 
We train a \emph{lifelong} model and a typical VAE \emph{baseline}
without catastrophic forgetting mitigation strategies and evaluate the
final model's generations in Figure \ref{celeba_exp}. As visually
demonstrated in Figure \ref{celeba_exp}-\emph{Left}, the \emph{lifelong} model is
able to generate instances from all of the previous distributions,
however the \emph{baseline} model catastrophically forgets (Figure
\ref{celeba_exp}-\emph{Right}) and only
generates samples from the \emph{eye-glasses} distribution. This is also
reinforced by the \emph{log}-Frechet distance shown in Figure
\ref{ca_fid}.

We also evaluate the wall-clock time in seconds (Table
\ref{wall_clock}) for the \emph{lifelong} model and the
\emph{baseline-vae} for the 44,218 samples of the \emph{male}
distribution. We observe that the \emph{lifelong} model does not add a significant
overhead, especially since the \emph{baseline-vae} undergoes
catastrophic forgetting (Figure \ref{celeba_exp} \emph{Right}) and completely fails to
generate samples from previous distributions. Note that we present the number of
parameters and other \emph{detailed model information} in our code and
Appendix \ref{arch}.

\section{Ablation Studies}
In this section we independently validate the benefit of each of the
newly introduced components to the learning objective proposed in
Section \ref{full_learning_objective}. In Experiment
\ref{experiment0} we demonstrate the benefit of the
discrete-continuous posterior factorization introduced in Section
\ref{sampling}. 
Then in Experiment \ref{ablation}, we validate the necessity of the information restricting
regularizer (Section \ref{information_regularizer}) and posterior
consistency regularizer (Section \ref{regularizer}).

\subsection{Linear Separability of Discrete and Continuous Posterior} \label{experiment0}

\begin{figure}[H]
\begin{center}
\minipage{0.52\textwidth}
\includegraphics[width=\linewidth]{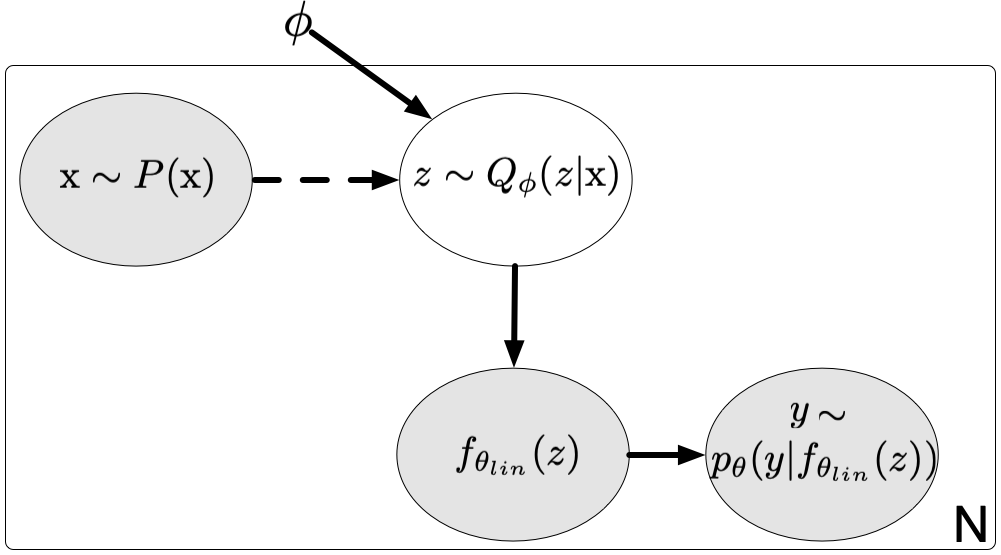}
\endminipage\hfill
\minipage{0.48\textwidth}
\includegraphics[width=\linewidth]{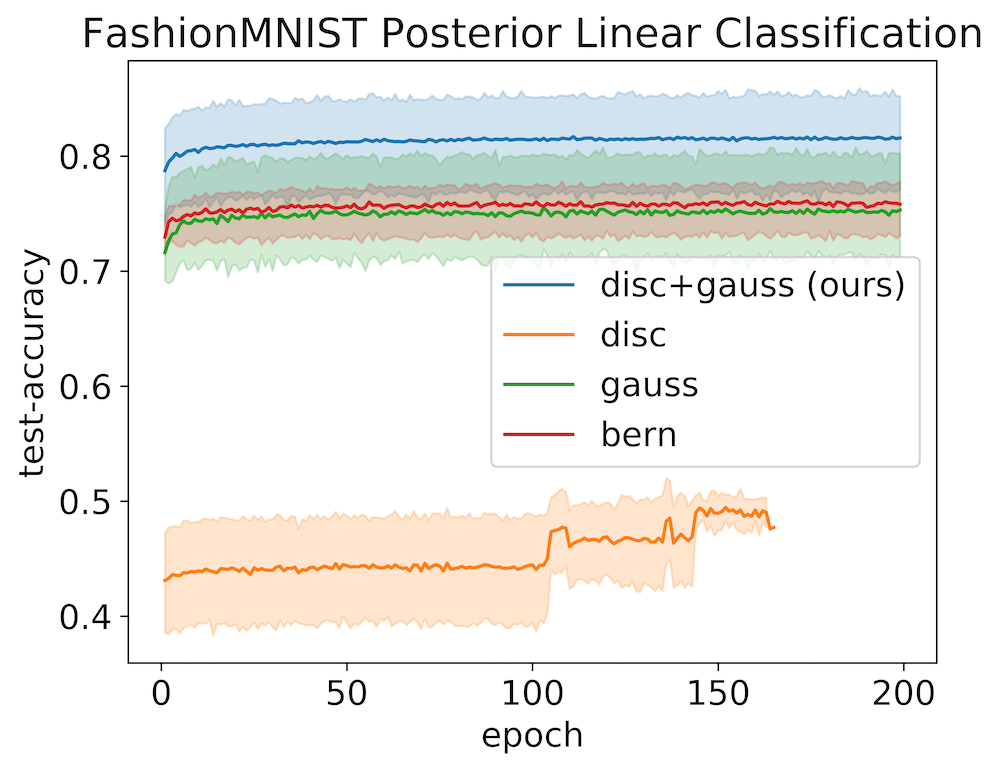}
\endminipage%
\caption{\emph{Left:} Graphical model depicting classification using
  pretrained VAE, coupled with a linear classifier,
  $f_{\bm{\theta}_{lin}} : \bm{z} \mapsto \bm{y}$. \emph{Right:}
  Linear classifier accuracy on the Fashion MNIST test set for a
  varying range of latent dimensions, $| \bm{z} | \in [32, 64, 128,
  256, 512, 1024]$ and distributions.} \label{linear_classifier}
\end{center}
\end{figure}
In order to validate that the (independent) discrete and
continuous latent variable posterior, $Q_{\bm{\phi}}(\bm{z}_d,
\bm{z}_c | \bm{\mathrm{x}})$, aids in learning a better representation, we classify the encoded
posterior sample using a simple linear classifier
$f_{\bm{\theta}_{lin}} : \bm{z} \mapsto \bm{y}$, where $\bm{y}$
corresponds to the categorical class prediction. Higher
(linear) classification accuracies demonstrate that the the VAE is able to
learn a more linearly separable representation. Since the latent
representation of VAEs are typically used in auxiliary tasks, learning
such a representation is useful in downstream tasks. This is
a standard method to measure posterior separability and is used in
methods such as Associative Compression Networks
\cite{graves2018associative}.

We use the standard training set of FashionMNIST
\cite{xiao2017/online} (60,000 samples) to train a standard VAE with a discrete only (\emph{disc}) posterior,
an isotropic-gaussian only (\emph{gauss}) posterior, a bernoulli only
(\emph{bern}) posterior and finally the proposed
independent discrete and continuous (\emph{disc+gauss}) posterior presented in Section
\ref{sampling}. For each different posterior reparameterization, we train a set of VAEs with varying latent
dimensions, $| \bm{z} | \in [32, 64, 128, 256, 512, 1024]$. In the case of the
\emph{disc+gauss} model we fix the discrete dimension, $|\bm{z}_d| = 10$ and vary the isotropic-gaussian dimension to match the total
required dimension. After training each VAE, we proceed to use the same
training data to train a linear classifier on the encoded posterior
sample, $\bm{z} \sim Q_{\bm{\phi}}(\bm{z} | \bm{\mathrm{x}})$.

In Figure \ref{linear_classifier} we present the mean and standard
deviation linear test classification accuracies of each set of the different
experiments. As expected, the discrete only (\emph{disc}) posterior
performs poorly due to the strong restriction of mapping an entire input
sample to a single one-hot vector. The isotropic-gaussian
(\emph{gauss}) and bernoulli (\emph{bern}) only models provide a strong
baseline, but the combination of isotropic-gaussian and discrete
posteriors (\emph{disc+gauss}) performs much better, reaching an
upper-bound (linear) test-classification accuracy of \textbf{87.1\%}. This validates that the decoupling of
latent represention presented in Section \ref{sampling} aids in
learning a more meaningful, separable posterior.

\subsection{Validating the Mutual Information and Posterior Consistency Regularizers.}\label{ablation}
\begin{figure*}[ht]
\begin{center}
  \scalebox{1.0}{
\minipage{0.72\textwidth}
\includegraphics[width=\linewidth]{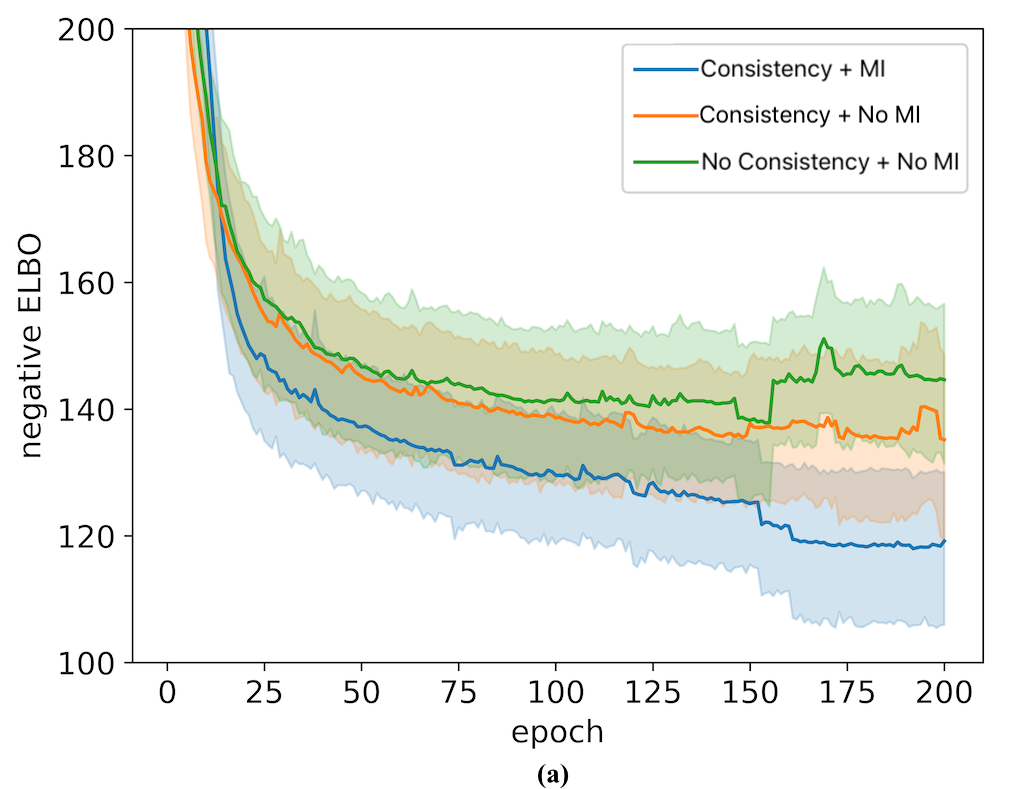}
    \endminipage\hfill
  }\scalebox{1.0}{
\minipage{0.28\textwidth}
\includegraphics[width=\linewidth]{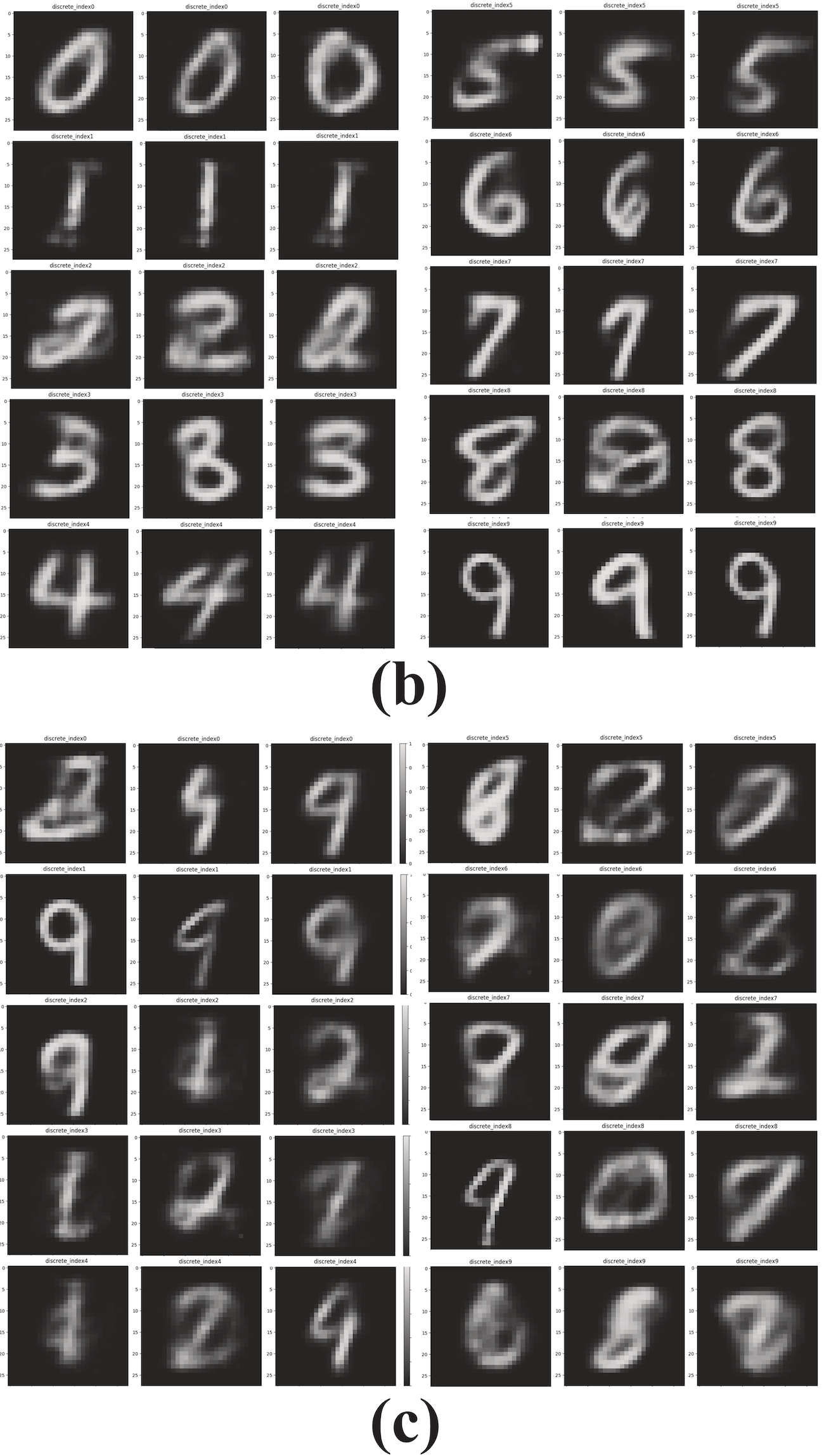}
\endminipage%
}
\end{center}%
\caption{MNIST Ablation: (a) negative test ELBO. (b) Sequentially
  generated samples by setting $\bm{z}_d$ and sampling $\bm{z}_c \sim
  \mathcal{N}(0, 1)$ (Section \ref{sampling}) with \emph{consistency + mutual
  information} (MI). (c) Sequentially
  generated samples with \emph{no consistency + no mutual
  information} (MI).}\label{fig_ablation}
\end{figure*}%
In order to independently evaluate the benefit of our proposed
Bayesian update regularizer (Section \ref{regularizer}) and the
mutual information regularizer proposed in (Section \ref{sampling}) we
perform an ablation study 
using the MNIST data flow sequence from Figure \ref{data_flow_fig}. 
We evaluate three scenarios: 1) with posterior consistency and mutual information regularizers, 2) only
posterior consistency and 3) without both regularizers.
We observe that both components are necessary in order to generate high
quality samples as evidenced by the negative test ELBO in Figure
\ref{fig_ablation}-(a) and the corresponding generations in Figure
\ref{fig_ablation}-(b-c). The generations produced
 without the information gain regularizer and consistency in Figure
 \ref{fig_ablation}-(c) are blurry. We attribute this to: 1) uniformly sampling the discrete component is not
guaranteed to generate samples representative samples from
$P_{<i}(\bm{\mathrm{x}})$ and 2) the decoder,
$P_{\bm{\Theta}}(\bm{\mathrm{x}} | \bm{z}_d, \bm{z}_c)$, relays more
information through the continuous component, $P_{\bm{\Theta}}(\bm{\mathrm{x}} | \bm{z}_d, \bm{z}_c) =
P_{\bm{\Theta}}(\bm{\mathrm{x}} | \bm{z}_c)$, causing
catastrophic forgetting and posterior collapse \cite{alemi2018fixing}.


\section{Limitations}\label{limitations}
While LGM presents strong performance, it fails to completely
solve the problem of lifelong generative modeling and we see a slow degradation in model performance over
time. We attribute this mainly to the problem of poor VAE generations
that compound upon each other (also discussed below). In addition,
there are a few poignant issues that need to be resolved in order to
achieve an optimal (in terms of non-degrading Frechet distance / -ELBO) unsupervised generative lifelong learner:

\textbf{Distribution Boundary Evaluation}: The standard assumption in current lifelong / continual learning
approaches \cite{nguyen2018variational,
  zenke2017continual,shin2017continual,kamra2017deep,kirkpatrick2017overcoming,rusu2016progressive}
is to use known, fixed distributions instead of learning the
distribution transition boundaries. 
For the purposes of this work, we focus on the
accumulation of distributions (in an unsupervised way), rather than
introduce an additional level of indirection through the incorporation
of anomaly detection methods that aid in detecting distributional boundaries.

\textbf{Blurry VAE Generations}: VAEs are known to generate images that are blurry in contrast to GAN
based methods. This has been attributed to the fact that VAEs don't learn the true posterior
and make a simplistic assumption regarding the reconstruction distribution
$P_{\bm{\theta}}(\bm{\mathrm{x}} | \bm{z})$
\cite{alemi2018fixing,rainforth2018tighter}. While there exist methods
such as ALI \cite{dumoulin2016adversarially} and BiGAN
\cite{donahue2016adversarial}, that learn a posterior distribution
within the GAN framework, recent work
has shown that adversarial methods fail to accurately match
posterior-prior distribution ratios in large dimensions \cite{rosca2018distribution}.

\textbf{Memory}: In order to scale to a truly lifelong setting, we
posit that a learning algorithm needs a global pool of memory that can
be decoupled from the learning algorithm itself. This decoupling would
also allow for a principled mechanism for parameter transfer between
sequentially learnt models as well a centralized location for
compressing non-essential historical data. Recent work such as the Kanerva Machine
\cite{wu2018kanerva} and its extensions \cite{wu2018learning} provide
a principled way to do this in the VAE setting.

\section{Conclusion}

In this work we propose a novel method for learning generative models
over a lifelong setting.
The principal assumption for the data is that they are generated by multiple distributions and
presented to the learner in a sequential manner. 
A key \emph{limitation} for the learning process is that the method
\emph{has no access to any of the old data and that it shall distill
  all the necessary information into a single final model}.
The proposed method is based on a dual student-teacher architecture where the
teacher's role is to preserve the past knowledge and aid the student in future
learning. We argue for and augment the standard VAE's ELBO objective by terms
helping the teacher-student knowledge transfer. We demonstrate the benefits this augmented objective brings to the lifelong
learning setting using a series of
experiments. The architecture, combined with the proposed regularizers,
aid in mitigating the effects of catastrophic interference by supporting the retention of previously learned knowledge.

%% file: appendix.tex
\section{Appendix}\label{appendix}

\subsection{Understanding the Consistency
  Regularizer}\label{continuous_regularizer}

The analytical derivations of the consistency
regularizer show that the regularizer can be interpreted as an a transformation
of the standard VAE regularizer. In the case of an isotropic gaussian posterior,
the proposed regularizer scales the mean and variance of the student
posterior by the variance of the teacher \ref{contreg} and adds an extra
'volume' term. This interpretation of
the consistency regularizer shows that the proposed regularizer preserves the
same learning objective as that of the standard VAE. Below we present the
analytical form of the consistency regularizer with categorical and
isotropic gaussian posteriors:


\begin{proof} \label{contreg}
  We assume the learnt posterior of
  the teacher is parameterized by a centered, isotropic gaussian with
  $\bm{\Phi} = [\bm{\mu^E = \bm{0}}, \bm{\Sigma^E} = \text{diag}(\bm{\sigma^{E^2}})]$
  and the posterior of our student by a non-centered isotropic gaussian with $\bm{\phi} =
  [\bm{\mu^S}, \bm{\Sigma^S} = \text{diag}(\bm{\sigma^{S2}})]$, then
  \begin{center}
\begin{adjustbox}{width=1.0\columnwidth}
  \parbox{1.0\linewidth}{%
  \begin{align}
    \begin{split}
      KL(Q_{\bm{\phi}}(\bm{z}|\bm{x}) || Q_{\bm{\Phi}}(\bm{z} | \bm{x})) &= 0.5\bigg[tr(\bm{\Sigma}^{E^{-1}} \bm{\Sigma^S}) + (\bm{\mu^E} - \bm{\mu^S})^ T\bm{\Sigma}^{E^{-1}}(\bm{\mu^E} - \bm{\mu^S}) - F + log \bigg( \frac{|\bm{\Sigma^E}|}{|\bm{\Sigma^S}|} \bigg)\bigg] \\
      &= 0.5\sum_{j=1}^{F}\bigg[\frac{1}{\sigma^{E2}(j)}(\sigma^{S2}(j) + \mu^{S2}(j)) - 1 + log\  \sigma^{E2}(j) - log\  \sigma^{S2}(j)\bigg] \\
      &= KL(Q_{\bm{\phi^{*}}}(\bm{z}|\bm{x}) || \mathcal{N}(0, \bm{I})) - log\ |\bm{\Sigma}^E|
      \end{split}
  \end{align}
  Via a reparameterization of the student's parameters:
  \begin{align}
    \begin{split}
    \bm{\phi}^{*} &= [\bm{\mu}^{S*}, \bm{\sigma}^{{S*2}}] \\
    \bm{\mu}^{S*} = \frac{\mu^{S}(j)}{\sigma^{E2}(j)} &; \bm{\sigma}^{S*2} = \frac{\sigma^{S2}(j)}{\sigma^{E2}(j)}
    \end{split}
  \end{align}
}
\end{adjustbox}
\end{center}
\end{proof}\label{cor1}

It is also interesting to note that our posterior regularizer becomes the prior if:
$$lim_{\bm{\sigma}^{E^2} \mapsto
    1}KL(Q_{\bm{\phi}}(\bm{z}|\bm{x}) || Q_{\bm{\Phi}}(\bm{z} | \bm{x}))
  = KL(Q_{\bm{\phi}}(\bm{z}|\bm{x}) || \mathcal{N}(0, \bm{I}))$$

\begin{proof}
  We parameterize the learnt posterior of the teacher by $\Phi_i =
  \frac{\exp({p_i^E})}{\sum_{i=1}^J \exp({p_i^E})}$ and the posterior of the
  student by $\phi_i = \frac{\exp({p_i^S})}{\sum_{i=1}^J \exp({p_i^S})}$. We
  also redefine the normalizing constants as $c^E = \sum_{i=1}^J \exp({p_i^E})$
  and $c^S = \sum_{i=1}^J \exp({p_i^S})$ for the teacher and student models
  respectively.
  The KL divergence from the ELBO can now be re-written as:
  \begin{align}
    \begin{split}
      KL(Q_{\bm{\phi}}(\bm{z}_d|\bm{x}) || Q_{\bm{\Phi}}(\bm{z}_d | \bm{x})) &= \mathlarger{\sum_{i = 1}^{J}} \frac{\exp({p_i^S})}{c^S} log\ \bigg( \frac{\exp({p_i^S})}{c^S} \frac{c^E}{\exp({p_i^E})} \bigg) \\
      &= H(\bm{p}^S, \bm{p}^S - \bm{p}^E) = -H(\bm{p}^s) + H(\bm{p}^S, \bm{p}^E)
    \end{split}
  \end{align} \label{cor0}
  where $H(\_)$ is the entropy operator and $H(\_,\_)$ is the
  cross-entropy operator.
\end{proof}

\subsection{Reconstruction Regularizer}\label{ll_regularizer}

\begin{figure}[H]
\minipage{0.5\textwidth}
\includegraphics[width=\linewidth]{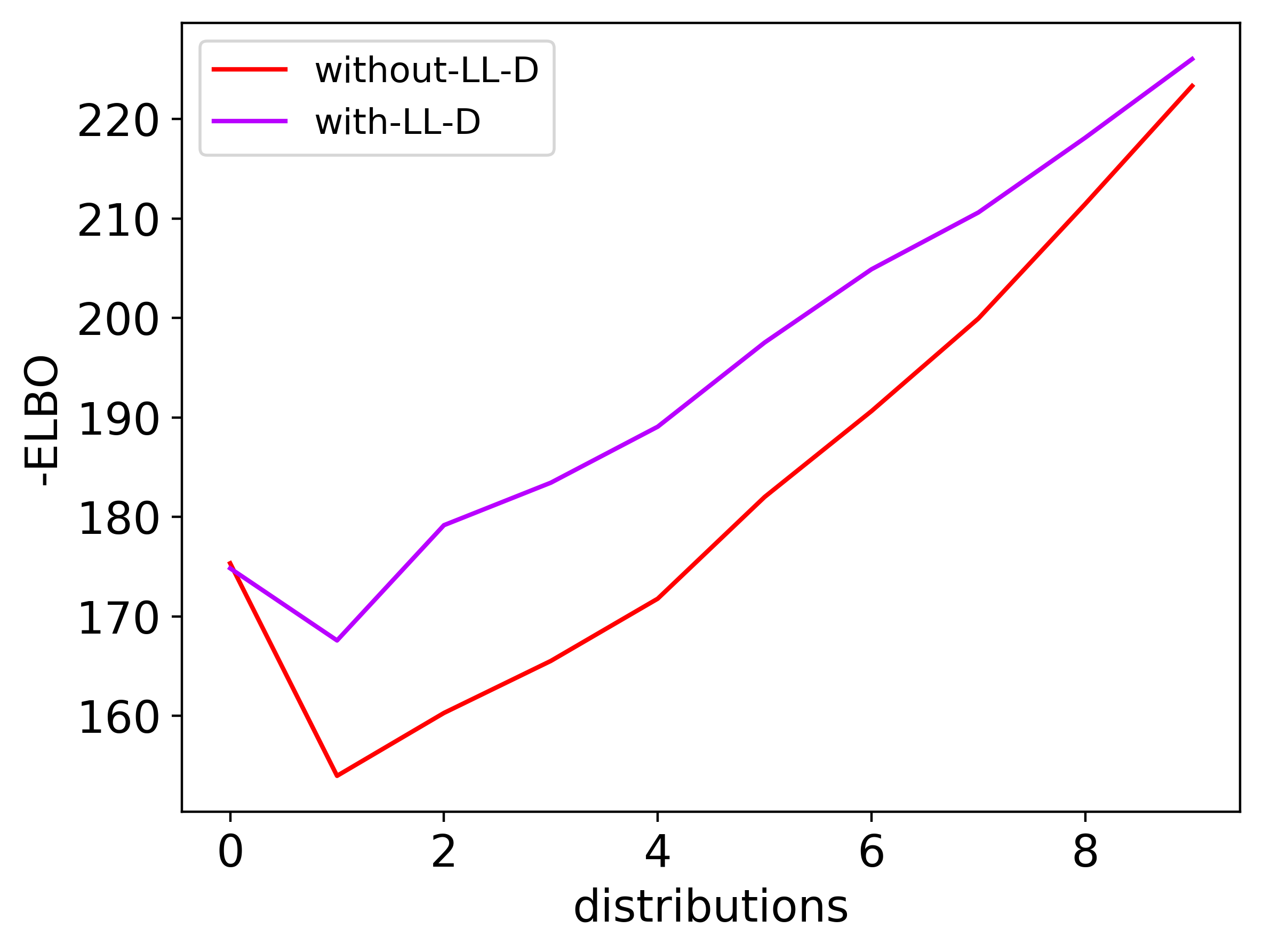}
  \caption{Fashion Negative Test ELBO}\label{ll_regularizer_fig1}
\endminipage\hfill
\minipage{0.5\textwidth}
\includegraphics[width=\linewidth]{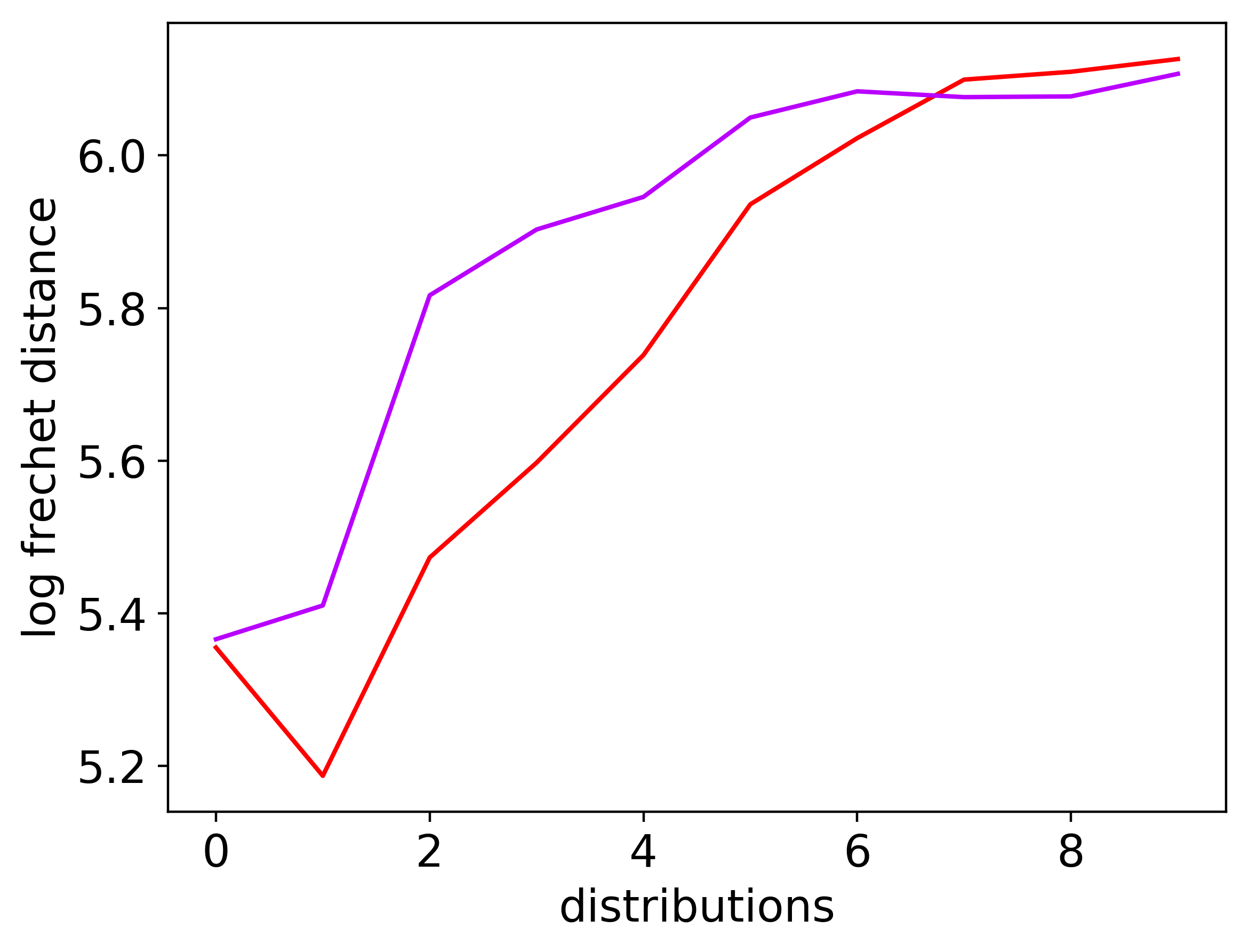}
  \caption{Fashion Log-Frechet Distance }\label{ll_regularizer_fig2}
\endminipage\hfill
\end{figure}

While it is possible to constrain the reconstruction/decoder term of the VAE in
a similar manner to the consistency posterior-regularizer,
i.e: $KL[P_{\bm{\theta}}(\hat{\bm{\mathrm{x}}}| \bm{z}) ||
P_{\bm{\Theta}}(\hat{\bm{\mathrm{x}}} | \bm{z})]$, doing so diminishes
model performance. We hypothesize that this is due to the
fact that this regularizer contradicts the objective of the reconstruction term
$P_{\bm{\theta}}(\bm{\mathrm{x}}| \bm{z})$ in the ELBO which
already aims to minimize some metric between the input samples
$\bm{\mathrm{x}}$ and the
reconstructed samples $\hat{\bm{\mathrm{x}}}$; eg: if $P_{\bm{\theta}}(\bm{\mathrm{x}}|
\bm{z}) \sim \mathcal{N}(\bm{\mu}, \text{diag}[\bm{\sigma}])$, then the loss is proportional
to $|| \hat{\bm{\mathrm{x}}} - \bm{\mathrm{x}}||_2^2$, the standard L2
loss. Without the addition of this reconstruction cross-model
regularizer, the model is also provided with more flexibility in how it
reconstructs the output samples.

In order to quantify the this we duplicate the FashionMNIST Experiment
listed in the data flow definition in Figure \ref{data_flow_fig}. We
use a simpler model than the main experiments to validate this hypothesis.
We train two dense models (-D): one with just the
posterior consistency regularizer (\emph{without-LL-D}) and one with the
consistency and likelihood regularizer (\emph{with-LL-D}). We
observe the model performance drops (with respect to the Frechet
distance as well the test ELBO) in the case of the
\emph{with-LL-D} as demonstrated in Figures
\ref{ll_regularizer_fig1} and \ref{ll_regularizer_fig2}.


\subsection{Model Architecture}\label{arch}

We used two different architectures for our experiments.
When we use a dense network (-D) we used two layers of 512 units to map to the
latent representation and two layers of 512 to map back to the reconstruction
for the decoder. We used batch norm \cite{ioffe2015batch} and ELU activations for
all the layers barring the layer projecting into the latent representation and
the output layer. Note that while we used the same architecture for
EWC we observed a drastic negative effect when using batch norm and thus
dropped it's usage. The convolution architectures (-C) used the
architecture described below for the encoder and the decoder (where the decoder used
conv-transpose layers for upsampling).  The notation is
[OutputChannels, (filterX, filterY), stride]:
\begin{align}
  \begin{split}
  \text{Encoder: } &[32, (5, 5), 1] \mapsto \text{GN+ELU} \mapsto [64, (4,
4), 2] \mapsto  \text{GN+ELU} \mapsto [128, (4, 4), 1]
\mapsto  \\ &\text{GN+ELU}
\mapsto [256, (4, 4), 2]
  \mapsto  \text{GN+ELU} \mapsto
  [512, (1, 1), 1] \mapsto  \\ &\text{GN+ELU} \mapsto [512, (1, 1), 1] \\
  \text{Decoder: } &[256, (4, 4), 1] \mapsto \text{GN+ELU} \mapsto [128, (4,
4), 2] \mapsto  \text{GN+ELU} \mapsto [64, (4, 4), 1] \\
&\mapsto  \text{GN+ELU}
\mapsto [32, (4, 4), 2] \mapsto  \text{GN+ELU} \mapsto
[32, (5, 5), 1] \\ &\mapsto  \text{GN+ELU} \mapsto [\text{chans}, (1, 1), 1]
  \end{split}
\end{align}


\scalebox{0.8}{
\begin{tabular}{ c | c | c | c | c | c} \hline
  Method & Initial $\bm{z}_d$ dimension & Final $\bm{z}_d$ dimension
  & $\bm{z}_c$ dimension & \# initial parameters & \# final parameters
  \\ \hline
  EWC-D & 10 & 10 & 14 & 4,353,184 & 4,353,184 \\
naive-D & 10 & 10 & 14 & \textbf{1,089,830} & \textbf{1,089,830}
  \\
  batch-D & 10 & 10 & 14 & \textbf{1,089,830} & \textbf{1,089,830} \\
  batch-D & 10 & 10 & 14 & 2,179,661 & 2,179,661 \\
  lifelong-D & 1 & 10 & 14 & 2,165,311 & 2,179,661 \\ \hline
  EWC-C & 10 & 10 & 14 & 30,767,428 & 30,767,428 \\
  naive-C & 10 & 10 & 14 & \textbf{7,691,280} & \textbf{7,691,280}
  \\
  batch-C & 10 & 10 & 14 & \textbf{7,691,280} & \textbf{7,691,280} \\
  batch-C & 10 & 10 & 14 & 15,382,560 & 15,382,560 \\
  lifelong-C & 1 & 10 & 14 & 15,235,072 & 15,382,560 \\
\end{tabular}\label{hyperparams}}
\vskip 0.1in

The table above lists the number of parameters for each model
and architecture used in our experiments. The \emph{lifelong}
models initially start with a $\bm{z}_d$ of dimension 1 and at each
step we grow the representation by one dimension to accommodate the new
distribution (more info in Section \ref{expandablerep}). In contrast,
the baseline EWC models are provided with the full
representation throughout the learning process. EWC has double the
number of parameters because the computed diagonal fisher information
matrix which is the same dimensionality as the number of parameters. EWC also
neeeds the preservation of the teacher model $[\bm{\Phi},
\bm{\Theta}]$ to use in it's quadratic regularizer. Both the \emph{naive}
and \emph{batch} models have the fewest number of parameters as they do not
use a student-teacher framework and only use one model, however the
vanilla model has no protection against catastrophic interference and
the \emph{full} model is just used as an upper bound for performance.

We used Adam \cite{kingma2015adam} to optimize all of our
problems with a learning rate of 1e-4 or 1e-3. When we used weight transfer we
re-initialized the accumulated momentum vector of Adam as well as the aggregated
mean and variance of the batch norm layers.
The full architecture can be examined in our github
repository \cite{jramapuram_2018} and is provided under an MIT license.



\input{streaming}

\subsection{EWC Baselines: Comparing Conv \& Dense
  Networks}\label{ewc_baselines}
\begin{figure}[H]
  \includegraphics[width=\linewidth]{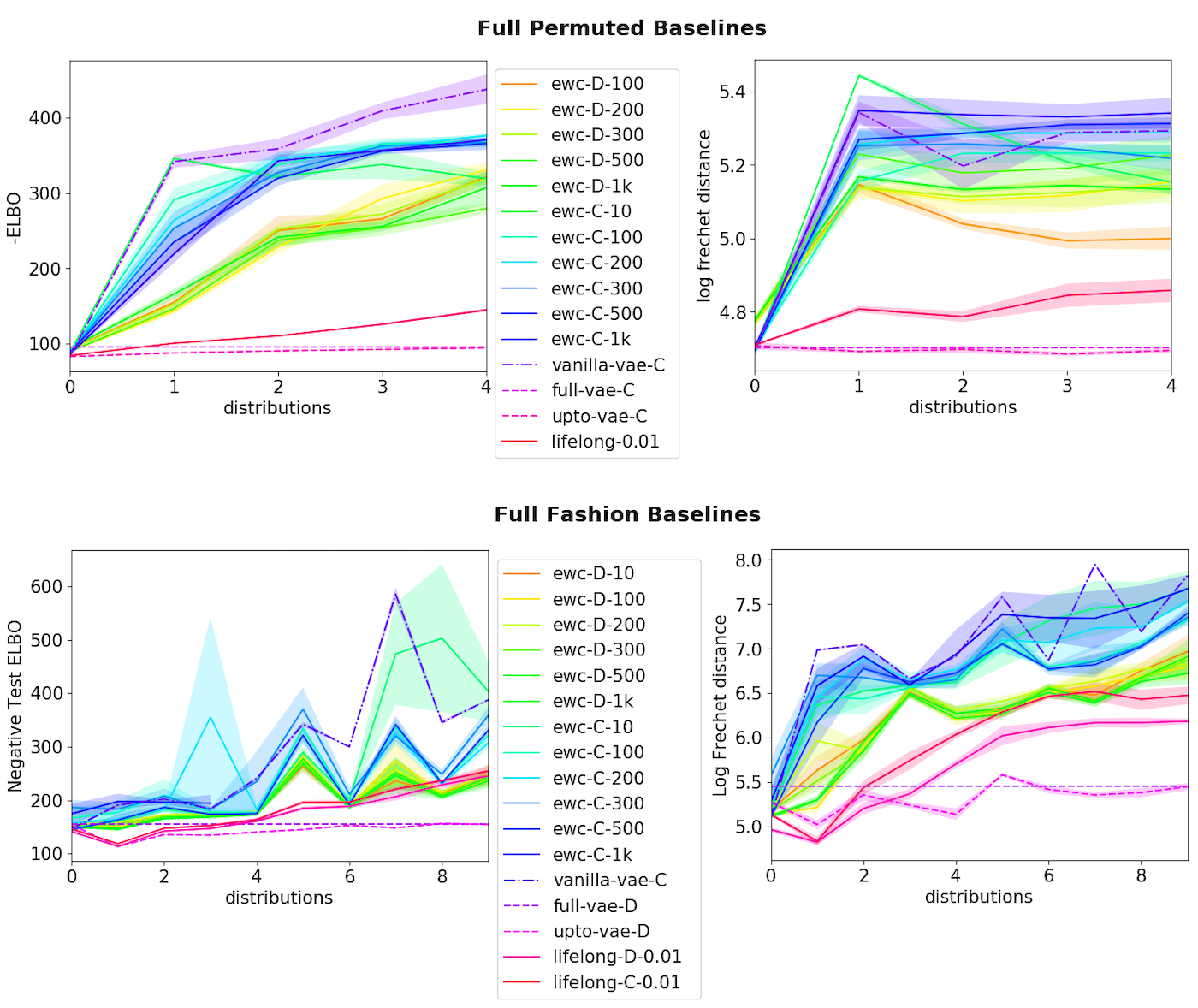}
\end{figure}\label{full_baselines}

We compared a whole range of EWC baselines and use the best
performing models few in our experiments. Listed in Figure \ref{full_baselines} are the full range of EWC baselines run
on the PermutedMNIST and FashionMNIST experiments. Recall that C / D describes whether
a model is convolutional or dense and the the number following  is the
hyperparameter for the EWC or Lifelong VAE.

\subsection{Gumbel Reparameterization}
Since we model our latent variable
as a combination of a discrete and a continuous distribution we also use the
Gumbel-Softmax reparameterization
\cite{maddison2016concrete,jang2016categorical}. The Gumbel-Softmax
reparameterization over logits [linear output of the last layer in
the encoder] $\bm{p} \in \mathcal{R}^{M}$ and an annealed
temperature parameter $\tau \in \mathcal{R}$ is
defined as:
\begin{align}
  \bm{z} &= softmax(\frac{{log(\bm{p}) + \bm{g}}}{\tau}); \bm{g} = -log(-log(\bm{u} \sim Unif(0, 1)))
\end{align}\label{gumbel}
$\bm{u} \in \mathcal{R}^{M}, \bm{g} \in \mathcal{R}^{M}$. As
the temperature parameter $\tau \mapsto 0$, $\bm{z}$
converges to a categorical. 

\subsection{Expandable Model Capacity and Representations}\label{expandablerep}
Multilayer neural networks with sigmoidal activations have a VC dimension bounded
between $O(\rho^2)$\cite{sontag1998vc} and
$O(\rho^4)$\cite{karpinski1997polynomial} where $\rho$ are the number of
parameters. A model that is able to consistently add new information should also
 be able to expand its VC dimension by adding new parameters over
time. Our formulation imposes no restrictions on the model architecture: i.e.
new layers can be added freely to the new student model.

In addition we also allow the dimensionality of $\bm{z}_d \in \mathcal{R}^J$,
our discrete latent representation to grow in order to accommodate new
distributions. This is possible because the KL divergence between two
categorical distributions of different sizes can be evaluated by simply zero
padding the teacher's smaller discrete distribution. Since we also transfer
weights between the teacher and the student model, we need to handle the case of
expanding latent representations appropriately. In the event that we add a new
distribution we copy all the weights besides the ones immediately surrounding
the projection into and out of the latent distribution. These surrounding
weights are reinitialized to their standard Glorot initializations
\cite{glorot2010understanding}.

%% file: streaming.tex
\subsection{Contrast to streaming / online methods}\label{diff_streaming}

Our method has similarities to streaming methods such as Streaming
Variational Bayes (SVB) \cite{DBLP:conf/nips/BroderickBWWJ13} and
Incremental Bayesian Clustering methods \cite{katakis2008incremental, gomes2008incremental} in that we
estimate and refine posteriors through time. In general this can be
done through the following Bayesian update rule that states that the
lastest posterior is proportional to the current likelihood times the previous
posterior:

\begin{align}
P(\bm{z} | \bm{\mathrm{X}}_1, ..., \bm{\mathrm{X}}_t)
\propto P(\bm{\mathrm{X}}_t | \bm{z})P(\bm{z}|\bm{\mathrm{X}}_1,..., \bm{\mathrm{X}}_{t-1}) \label{bayes_update_rule}
\end{align}
SVB computes the intractable posterior, $P(\bm{z} | \bm{\mathrm{X}}_1,
 ..., \bm{\mathrm{X}}_t)$, utilizing an approximation,
$\mathcal{A}_t$, that accepts as input the current dataset, $\bm{\mathrm{X}}_t$, along
with the previous posterior $\mathcal{A}_{t-1}$ :

\begin{align}
  P(\bm{z} | \bm{\mathrm{X}}_1, ..., \bm{\mathrm{X}}_t)
  \approx \mathcal{A}_t(\bm{\mathrm{X}}_t, \mathcal{A}_{t-1})
\end{align}

The first posterior input ($\mathcal{A}_{t=0}$) to the approximating function is
the prior $P(\bm{z})$. The objective of SVB and other streaming methods is to model the posterior
of the currently observed data in the best possible manner. Our setting differs
from this in that we want to retain information from \emph{all previously observed
distributions} (sometimes called a knowledge store \cite{thrun1995lifelong}). This can be useful in scenarios where a distribution
is seen once, but only used much later down the road. Rather than creating a posterior update rule, we
recompute the posterior via Equation \ref{bayes_update_rule}, leveraging the
fact that we can re-generate $\bm{\mathrm{X}}_{<t} \approx
\bm{\hat{\mathrm{X}}}_{<t}$ through the generative process. This
allows us to recompute a more appropriate posterior re-using all of the
(generated) data, rather than using the previously computed
(approximate) posterior $\mathcal{A}_{t-1}$:
\begin{align}
P(\bm{z} | \bm{\mathrm{X}}_1, \bm{\mathrm{X}}_2, ..., \bm{\mathrm{X}}_t)
  \propto P(\bm{\mathrm{X}}_t | \bm{z})P(\bm{z}|\bm{\hat{\mathrm{X}}}_1,
   ..., \bm{\hat{\mathrm{X}}}_{t-1}) \label{posterior_update}
\end{align}

Coupling this generative replay strategy with the Bayesian update regularizer
introduced in Section \ref{regularizer}, we demonstrate that not only
do we learn an updated poster as in Equation
\ref{posterior_update}, but also allow for a natural transfer of
information between sequentially learnt models: a fundamental tenant
of lifelong learning \cite{thrun1995lifelong,thrun1995lifelong2}.

Finally, another key difference between lifelong learning and online
methods is that lifelong learning aims to learn from a sequence of
\emph{tentatively different} \cite{chen2016lifelong}
tasks while still retaining and accumulating knowledge; online
learning generally
assumes that the true underlying distribution comes from a single
distribution  \cite{bottou1998online}. There are some exceptions to this where online learning
is applied to the problem of domain adaptation, eg:
\cite{jain2011online, katakis2008incremental}. 

